\documentclass[final,3p,times,twocolumn]{elsarticle}

\usepackage{amsmath,amssymb,amsfonts}
\usepackage{graphicx}
\usepackage{textcomp}
\usepackage{xcolor}
\usepackage{ulem}
\usepackage{float}
\usepackage{amssymb}
\usepackage{color}
\usepackage[pagewise]{lineno}
\usepackage{mdwlist}
\usepackage{algorithm}
\usepackage{algorithmicx}
\usepackage{algpseudocode}
\usepackage{amsmath}
\algrenewcommand\textproc{}
\usepackage[colorlinks]{hyperref}
\usepackage{float}
\usepackage{subfigure}

\usepackage{url}
\usepackage{multirow}

\usepackage{balance}
\usepackage{pifont}
\usepackage{amsmath}
\usepackage{array}

\usepackage[pagewise]{lineno}
\usepackage{amsmath}

\usepackage{mathrsfs}
\usepackage{tikz}

\usepackage{threeparttable}
\usepackage{supertabular,booktabs}
\usepackage{rotating}

\usepackage{supertabular}
\newcommand{\etal}{\textit{et al.}}
\newcommand{\ie}{\textit{i.e.}}
\newcommand{\eg}{\textit{e.g.}}

\begin{document}

\begin{frontmatter}



\title{Feature refinement: An expression-specific feature
learning and fusion method for micro-expression
recognition}

 \author[JS]{Ling Zhou}
 \ead{2111808003@stmail.ujs.edu.cn}
  \author[JS]{Qirong Mao}
 \ead{mao\_qr@ujs.edu.cn}
 \author[XH,OULU]{Xiaohua Huang\corref{cor1}}
 \ead{xiaohuahwang@gmail.com}
   \author[FF]{Feifei Zhang}
 \ead{feifei.zhang@ia.ac.cn}
 \author[ZH]{Zhihong Zhang}
 \ead{zhihong@xmu.edu.cn}

 \address[JS]{School of Computer Science and Communication Engineering, Jiangsu University, Zhenjiang, Jiangsu 212013, P.R. China}
  \address[XH]{School of Computer Engineering, Nanjing Institute of Technology, China}
  \address[FF]{National Laboratory of Pattern Recognition, Institute of Automation, Chinese Academy of Sciences, Beijing 100190, China}
  \address[ZH]{Xiamen University, Xiamen, China}
  \address[OULU]{Center for Machine Vision and Signal Analysis, University of Oulu, Finland}
 \cortext[cor1]{Corresponding author.}

\begin{abstract}

Micro-Expression Recognition has become challenging, as it is extremely difficult to extract the subtle facial changes of micro-expressions. Recently, several approaches proposed several expression-shared features algorithms for micro-expression recognition. However, they do not reveal the specific discriminative characteristics, which lead to sub-optimal performance. This paper proposes a novel Feature Refinement (\textbf{FR}) with expression-specific feature learning and fusion for micro-expression recognition.  It aims to obtain salient and discriminative features for specific expressions and also predict expression by fusing the expression-specific features. FR consists of an expression proposal module with attention mechanism and a classification branch. First, an inception module is designed based on optical flow to obtain expression-shared features. Second, in order to extract salient and discriminative features for specific expression, expression-shared features are fed into an expression proposal module with attention factors and proposal loss.
 Last,  in the classification branch,  labels of categories are predicted by a fusion of the expression-specific features.
Experiments on three publicly available databases validate the effectiveness of FR under different protocol.  Results on public benchmarks demonstrate that our FR provides salient and discriminative information for micro-expression recognition. The results also show our FR achieves better or competitive performance with the existing state-of-the-art methods on micro-expression recognition.
\end{abstract}


\end{frontmatter}


\section{Introduction}
\label{intro}
Micro-expression, a very brief and involuntary form of facial expressions occurring when people want to conceal one's true feelings, usually lasts between 0.04s to 0.2s \cite{Ekm2009}.
Automatic micro-expression analysis involves micro-expression spotting and micro-expression recognition (\textbf{MER}) \cite{Oh2018a}. Micro-expression spotting aims to automatically detect the temporal interval (from onset to offset) of a micro-movement in a sequence of video frames, also including the apex frame spotting, while MER refers to the classification task of identifying the micro-expression involved in the well-segmented video from onset to offset \cite{Oh2018a, Patel2015}. Our focus in this work is micro-expression recognition.

As an essential way of human emotional behavior understanding, MER has attracted increasing attention in human-centered computing in the past decades. Its potential applications~\eg~in police case diagnosis and psychoanalysis \cite{Ekman1969,Ekm2009} make it a core component in the next generation of computer system,  in which a natural human machine interface enables the user to account for subtle appearance changes of human faces, to reveal the hidden emotions \cite{Michael2010,Ekman2009} of humans, and to help understanding people's deceitful behaviors.

Although advances have been made, due to complex factors,~\eg~the subtle changes of micro-expressions, it is extremely difficult for MER to achieve superiority performance. Amongst, one critical research issue is how to extract salient and discriminative features from micro-expression. Currently, amount of feature descriptor methods have been proposed. They are commonly categorized into handcrafted features and deep learning features. However, these existing methods characterize the discrimination of micro-expressions inefficiently. On the other hand, it is time-consuming to design handcrafted feature~\cite{Xu2017} and manually adjust the optimal parameters~\cite{Lu2014}. 

To address these issues, we  proposed a simple yet efficient method, termed as Feature Refinement (\textbf{FR}), is proposed to extract the expression-specific feature. The FR consists of three feature refinement stages: expression-shared feature learning, expression-specific feature distilling, and  expression-specific feature fusion. In the first stage, a shallow two-stream Inception network with Inception blocks is designed to capture global and local information of optical flows for expression-shared feature learning. In the expression-specific feature distilling stage, based on expression-shared features, the proposal module with attention mechanism and proposal loss is proposed to distill expression-specific features for obtaining salient and discriminative features. The constraint of expression-specific objective function is designed to lead separate and different feature mapping. In the last stage, the element-wise sum function is used to fuse the separate expression-specific features for modeling expression-refined features under the deep network for  expression categories prediction. The fusion of expression-specific feature can boost the feature learning. These three stages constitute the whole process of feature refinement. Across these three stages of feature learning,  the novel deep network obtains the salient and discriminative representations for MER.

Overall, our contributions can be summarized as follows,
\begin{itemize}

\item We propose a deep learning based three-level  feature refinement architecture for MER. This architecture can effectively and automatically learn the salient and discriminative feature for micro-expression recognition by distilling expression-specific features and fusing these features to final expression-refined features for expression classification.

\item We propose a constructive but straightforward attention strategy and a simple proposal loss in expression proposal module for expression-specific feature learning.~Specifically, attention factors can capture the characteristics from the subtle changes of micro-expressions. On the other hand, penalization of the proposal loss can optimize the discrimination of features.

\item We extensively validate our FR on three benchmarks of MER. The experiment results sufficiently demonstrate the efficiency of expression-specific feature learning for micro-expression recognition, and provided the latest results for micro-expression recognition across three commonly available experimental protocols.

\end{itemize}

The rest of the paper is organized as follows. Section \uppercase\expandafter{\romannumeral2} introduces the related works. Section \uppercase\expandafter{\romannumeral3} presents our Inception-based feature learning algorithm in detail. Section \uppercase\expandafter{\romannumeral4} reports our experimental analysis. Section \uppercase\expandafter{\romannumeral5} discusses conclusions and future research directions. 
\section{Related Work}


\subsection{Handcrafted features}
The success of existing~traditional approaches in MER is attributed in good part to the quality of the handcrafted visual features representation. Normally, handcrafted features are fed into a supervised classifier~\eg~Support Vector Machines~\cite{Cortes1995} to train a  recognizer for the target expressions. The features are generally categorized into appearance-based and geometric-based features.

\subsubsection{Appearance-based features}
Local Binary Pattern from Three Orthogonal Planes (LBP-TOP) \cite{Zhao2007} is the most widely used appearance-based feature for micro-expression recognition. It combines the temporal features along with the spatial features from three orthogonal planes of the image sequence. As one of the earlier works in MER, LBP-TOP has been  widely used in micro-expression analysis. Due to its low computational complexity,  many LBP-TOP variants has been proposed,~\eg~LBP from three Mean Orthogonal Planes (LBP-MOP) \cite{Wang2015}, Spatiotemporal Completed Local Quantized Patterns (STCLQP) \cite{Huang2016}, and hierarchical spatiotemporal descriptors \cite{Zong2018a}. Wang~\etal~\cite{Wang2014} proposed a spatiotemporal descriptor utilizing six intersection points namely LBP with Six Intersection Points (LBP-SIP) to suppress redundant information in LBP-TOP and preserve more efficient computational complexity. Wang~\etal~\cite{Wang, Wang2015a} explored the influence of the color space for features extraction and extracted Tensor features from Tensor Independent Color Space (TICS), validating that color information can improve the recognition effect of LBP-TOP. Huang~\etal~\cite{Huang2015} proposed Spatiotemporal LBP with integral projection (STLBP-IP), which used facial shape information to improve recognition performance. Huang~\etal~\cite{Huang2019} further proposed discriminative spatiotemporal LBP with revisited integral projection (DiSTLBP-RIP) to reveal the discriminative information. Besides the LBP family,  3D Histograms of Oriented Gradients (3DHOG) \cite{Polikovsky2009, Polikovsky2013} was another appearance-based feature counting occurrences of gradient orientation in localized portions of the image sequence.

\subsubsection{Geometric-based features}
Geometric-based features aim to represent micro-expression samples by the aspect of face geometry,~\eg~shapes and location of facial landmarks. They can be categorized into optical flow based and texture variations based features. To estimate the apparent motion of objects, the optical flow method was mostly introduced to extract motion features in MER. To suppress facial identity appearance information on micro-expression, Lu~\etal~\cite{Lu2014} proposed Delaunay-based Temporal Coding Model (DTCM), which encoded the local temporal variation in each sub-region by Delaunay triangulation and standard deviation analysis.
Liu~\etal~\cite{Liu2016} proposed the Main Directional Mean Optical Flow (MDMO) to reduce the feature dimension by using optical flow in the main direction. The MDMO was least affected by the varied number of frames in the image sequence. Xu~\etal~\cite{Xu2017} designed Facial Dynamics Map (FDM) to suppress abnormal optical flow vectors which resulted from noise or illumination changes. Liong~\etal~\cite{Liong2018a} proposed Bi-Weighted Oriented Optical Flow (Bi-WOOF), which applied two schemes to weight the HOOF \cite{Chaudhry2009} descriptor locally and globally. The Bi-WOOF can obtain promising performance using only the onset and apex frame, increasing its effectiveness by a large margin.

Overall, handcrafted feature extraction~approach mostly relies on the manually designed extractor, which needs professional knowledge and complex parameter adjustment process. Meanwhile, each method suffers from poor generalization ability and robustness. Furthermore, due to the limited representation ability, engineered features may hardly handle the challenge of nonlinear feature warping caused by complicated situations,~\eg~under different environments.

\subsection{Deep learning features}
Recently, deep learning has been considered as an efficient way to learn feature representations. According to the different evaluation mechanisms, the existing methods were evaluated on the sole database, on Composite Database Evaluation protocol, and on Cross-database Micro-expression Recognition protocol, respectively.

\subsubsection{Features evaluated on the single database}

Evaluation on the single database means the training and testing samples are from the same micro-expression database.~Leave-One-Subject-Out (LOSO),  Leave-One-Video-Out (LOVO) or $K$-Fold rule is commonly used to evaluate deep learning model. For example, Kim~\etal~\cite{Kim2016} proposed a feature representation for the spatial information at different temporal states, which was based on the Long Short-Term Memory (LSTM) \cite{Hochreiter1997} recurrent neural network and only evaluated on a single dataset of  CASME~\uppercase\expandafter{\romannumeral2} \cite{Yan2014}.
 Peng~\etal~\cite{Peng2017} proposed Dual Temporal Scale Convolutional Neural Network (DTSCNN), which was evaluated on CASME \cite{Yan2013} and CASME~\uppercase\expandafter{\romannumeral2} databases. The DTSCNN was the first work in MER that utilized shallow two-stream neural network with inputs of optical-flow sequences.
 Nag~\etal~\cite{Nag2019} proposed an unified architecture for micro-expression spotting and recognition, in which spatial and temporal network  extracts time-contrasted features from the feature maps to contrast out subtle motions of micro-expressions. Wang~\etal~proposed transferring Long-term Convolutional Neural Network (TLCNN) \cite{Wang2018}  which utilized transfer learning from macro-expression to micro-expression database for MER. So far, there were many other deep learning features evaluated on the single database,~\eg~Spatiotemporal Recurrent Convolutional Networks (STRCN) \cite{Xia2020}, Three-stream 3D flow convolutional neural network \cite{Li2018}, Lateral Accretive Hybrid Network (LEARNet) \cite{Verma2019}, and 3D-Convolutional Neural Network method \cite{Reddy2019}.

\subsubsection{Features evaluated on Composite Database Evaluation protocol}

The Composite Database Evaluation (CDE) protocol \cite{See2019} means all the micro-expression databases are composited into one database and LOSO validation rule is used to evaluate the algorithm. As a deep learning work based on CDE protocol, Optical Flow Feature from Apex frame Network (OFF-ApexNet)~\cite{Gan2019} extracted optical flow features from the onset and apex frames of each video,  then learned features representation by feeding horizontal and vertical components of optical flows into a two-stream CNN network. Since the CNN network was shallow, it reduced the over-fitting caused by the scarcity of data in the micro-expression databases.

Very recently, more deep learning features were proposed \cite{Liong2019,Zhou2019,Liu2019,Quang2019} in MEGC 2019 \cite{See2019}. More specifically, Expression Magnification and Reduction (EMR) with adversarial training \cite{Liu2019} was a part-based deep neural network approach with adversarial training and expression magnification. With the special data augmentation strategy of expression magnification and reduction, EMR won the first place in MEGC 2019. Shallow Triple Stream Three-dimensional CNN (STSTNet) \cite{Liong2019} is an extended version of OFF-ApexNet. Besides the horizontal and vertical components of optical flows, STSTNet extracted the handcrafted feature named optical strain to learn more efficient features. By concentrating the three optical features to a single 3D image and then feeding the concentrated images to a shallow three-dimensional CNN,  STSTNet achieved state-of-the-art performance on CDE protocol among the methods without data augmentation. Dual-Inception~\cite{Zhou2019} was achieved by feeding the optical flow features extracted from the onset and mid-position frames into a designed two-stream Inception network. With data augmentation methods, Quang et al. \cite{Quang2019} applied  Capsule Networks (CapsuleNet) based on the apex frames to MER .

\subsubsection{Features evaluated on Cross-database Micro-expression Recognition protocol} Cross-database Micro-expression Recognition (CDMER) protocol means the samples of training and testing are selected from two different micro-expression databases~\cite{Zong2017,Yap2018,Zong2019}.
Based on the emotion classification, Zong~\etal~proposed domain generators approach for cross-database micro-expression recognition~\cite{Zong2018}. Holdout-database Evaluation (HDE) protocol \cite{Yap2018}, which is considered as a specific type of CDMER, was advocated in MEGC 2018 \cite{Yap2018}. This protocol aims to tackle the recognition of micro-expressions based on AU-centric objective classes rather than emotion classes. The two earliest works that introduced Deep Neural Network (DNN) on the HDE protocol were proposed by Peng~\etal~\cite{Peng2018} and Khor~\etal~\cite{Khor2018}. Specifically, Peng~\etal~used Resnet10 as a backbone and introduced transfer learning from macro-expression databases to learn micro-expression features~\cite{Peng2018}. Khor~\etal~adopted Enriched Long-term Recurrent Convolutional Network (ELRCN) to improve the recognition performance~\cite{Khor2018}, which contained the channel-wise for spatial enrichment and the feature-wise for temporal enrichment predicted the micro-expression by passing the feature vector through LSTM.

Although the aforementioned works have studied the problem of feature learning for MER, they primarily focused on learning expression-shared features from the input, ignoring resolving how to obtain salient and discriminative features. Due to the low intensity of micro-expression, generic feature learning lacks of revealing the intrinsic different characteristic of different micro-expressions. Expression-shared feature learning aims to learn identical feature space in the process of classification~\cite{Guo2019}. However, in~\cite{Guo2019}, the identical features for all categories may hardly lead to optimized performance. To solve these issues of expression-shared feature learning, this paper leverages expression-specific discriminant mapping features for MER. Specifically,  a straightforward feature refinement framework is proposed to learn salient and discriminative features for micro-expression recognition, which combines expression-specific features from expression-shared features by an expression proposal module with the attention mechanism.
\begin{figure*}[!t]
  \centering
  \includegraphics[width=\textwidth]{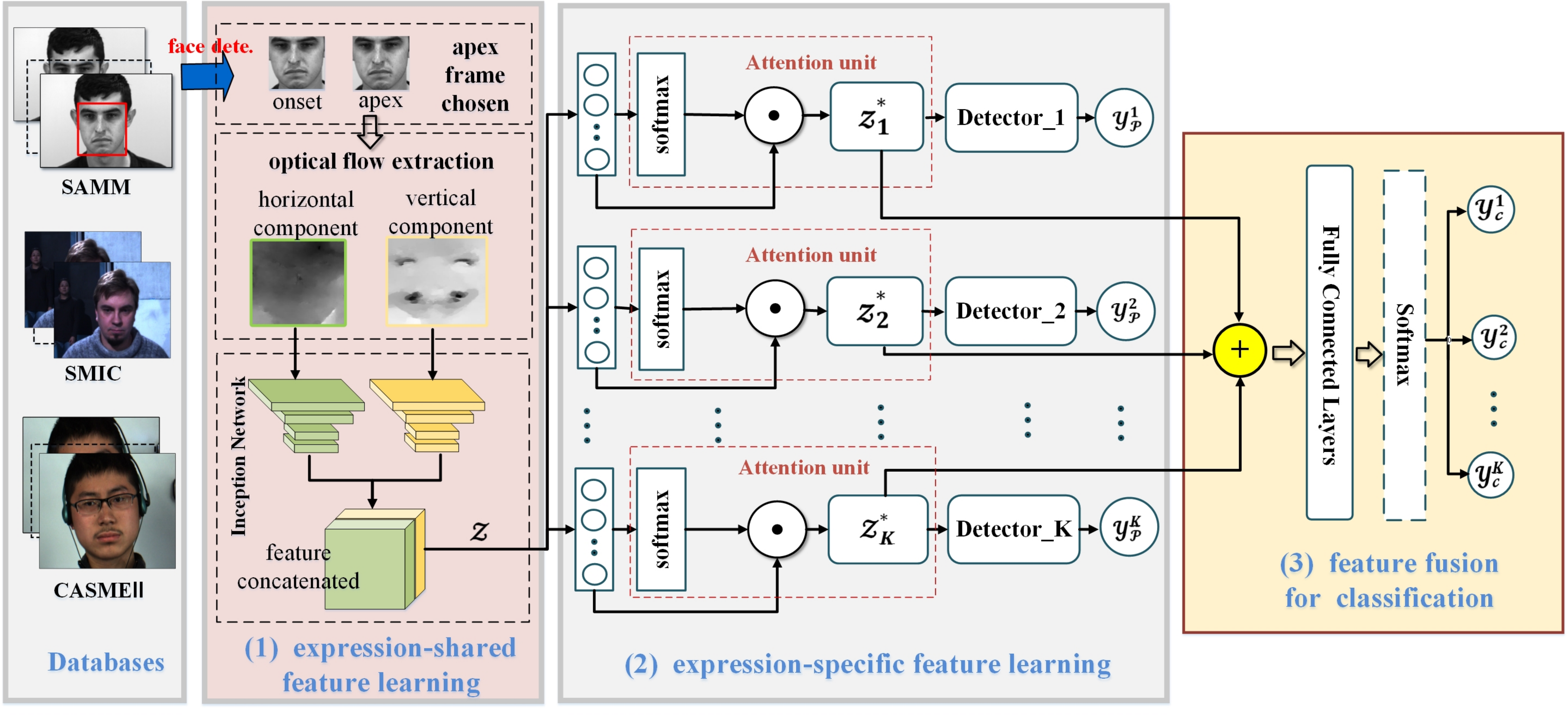}
  \caption{
  The architecture of the proposed FR for MER. The ``Databases'' module shows data from different micro-expression databases. The module ``expression-shared feature learning'' in the approach contains apex frame chosen, optical flow extraction and expression-shared feature learning in basic Inception network. The module ``expression-specific feature learning'' shows the expression-specific feature learning by using separate attention factors in the proposal module. The module ``feature fusion for classification'' illustrates that the expression-specific features are fused by element-wise sum function, and fused features are used to predict labels of categories.}
  \label{fr}
\end{figure*}

\section{Proposed Method}
\label{sec:ProposedMethod}
Fig.~\ref{fr} describes our proposed FR architecture. It leverages two-stream Inception network as a backbone for expression-shared feature learning, an expression proposal module with attention mechanism for expression-specific feature learning, and a classification module for label prediction by fused expression-refined features.

\subsection{Expression-shared feature learning}
As shown in Fig.~\ref{fr},  the expression-shared feature learning module consists of three critical components:  apex frame chosen,  optical flow extraction and two-stream Inception network.

Note that, as SMIC database~\cite{Li2013} doesn't supply the human-annotated apex frame, the apex frame spotting becomes very necessary. Several apex frame spotting algorithms have been proposed in recent years~\cite{Liong2015, Patel2015, Li2018b, Yan2014a,Peng,Zhou2019}. For example, the mid-position frame is straightforwardly chosen as the apex frame~\cite{Peng,Zhou2019}. Moreover, Liu~\etal~\cite{Liu2019} used motion difference to locate the apex frame. Quang~\etal~\cite{Quang2019} divided the face image into ten regions, and then computed the absolute pixel differences to find the apex frame. Considering a trade-off between efficiency and effectiveness, the inter-frame difference method (interframe-Diff) is presented to locate apex frame on SMIC database. The interframe-Diff defines the index of apex frames  $t_{apex}$ as follows:
\begin{equation}
t_{apex} = \arg \max{mean (I(x, y, t)-I(x, y, 0))},
\label{diff1}
\end{equation}
where $I(x,y,t)$ denotes the $t$-th frame of each sample, $mean()$ computes the mean absolute value of the pixel value difference between the onset and the $t$-th frames at $(x, y)$. 

As a motion information feature, the optical flow is extensively used by~\cite{Liong2019,  Gan2019, Liong} for micro-expression recognition. More specifically, the optical flow of each sample is extracted from the onset and apex frames, where onset and apex frames mean the frames with neutral-expression and the highest expression intensity, respectively. For FR, TV-L1 optical flow method~\cite{Zach2007} is utilized to obtain motion feature from the onset and apex frames of each micro-expression video. As shown in Fig.~\ref{fr}, for preserving more motion information, two optical flow components are extracted to represent the facial change along horizontal and vertical directions.

Furthermore, Considering horizontal and vertical components of optical flow, FR designs two-stream Inception network based on the Inception V1 block~\cite{Szegedy2015}, which complements with two-layer depths. Specifically, inception block is designed to capture both the global and local information of the optical component for feature learning. Distinguishing from traditional convolution with a fixed size of filters, Inception blocks parallelize filters of multiple sizes at the same level. Additionally, motivated by LeNet~\cite{Lecun1998}, the number of filters in the first and second layers is set at 6 and 16, respectively. Finally,  the flattened two sets of feature maps are concentrated into an expression-shared feature.
\begin{figure}
  \centering
  \includegraphics[width=0.35\textwidth]{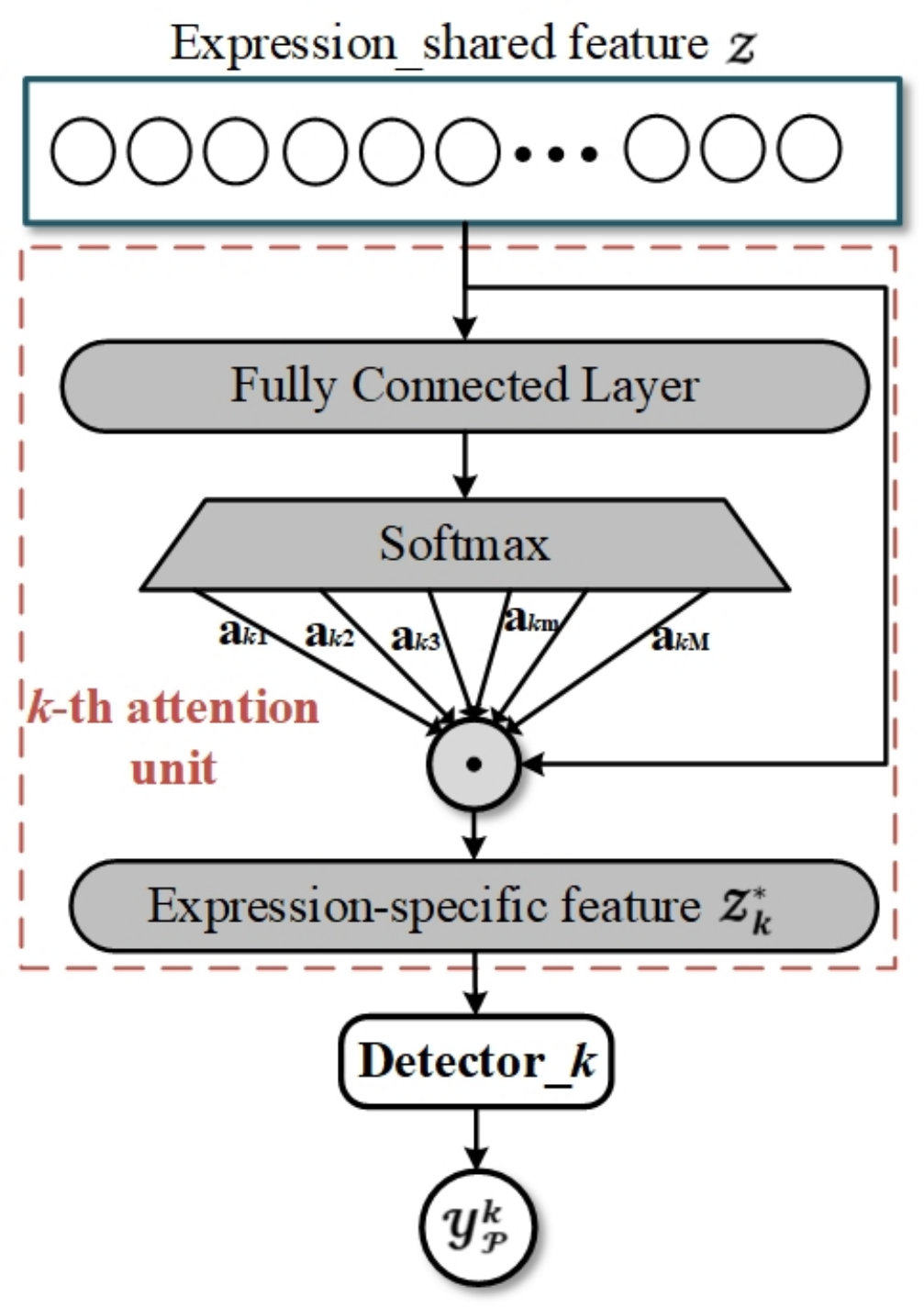}
  \caption{
  Attention unit in the proposal module for expression-specific feature learning.}
  \label{att_unit}
\end{figure}
\subsection{Expression-specific feature learning}

As illustrated in Fig.~\ref{fr}, based on the expression-shared feature, expression proposal module with the attention mechanism and proposal loss is introduced to learn the expression-specific features. Given a micro-expression sample $x$ and an expression-shared feature $z$, a proposal module with $K$ sub-branches is designed, where each of sub-branch learns a set of specific-feature for each micro-expression category (total $K$ categories). The proposal module becomes the core component of FR to obtain salient and discriminative features. The $Softmax$ attention and proposal loss are described as follows.

\textbf{Softmax attention:} Attention mechanism provides the flexibility to our model to learn $K$ expression-specific features from the same set of expression-shared feature. $K$ attention units are shown in the part (2) of Fig. \ref{fr}, and the detail of each attention unit is depicted in Fig. \ref{att_unit}. Specifically, the expression-shared feature $z$ is connected with $K$ fully connected layers separately. After activated by $Softmax$, we gain the attention weight of $z$ for each specific expression,
\begin{equation}
a_k= {softmax}_k(z),
\label{softmax}
\end{equation}
where $a_k$ is a vector that has the same dimension as $z$, and $k\leq K$. Then, the representation $z_k^*$ for each specific expression is given by:
\begin{equation}
z_k^*= a_k\ast z.
\label{softmax}
\end{equation}

\textbf{Proposal loss:} Besides the inducing of attention strategy, expression-specific feature learning is also constrained with a prososal loss which is averaged from $K$ expression-specific detection losses in the proposal module. Each expression-specific detector (`Detector$\_k$' in Fig.~\ref{fr}) is aligned with a feature vector $z_k^*$, which contains two fully connected layers with a $Sigmoid$~layer as the output. Thus, each sub-proposal branch is trained by optimizing the following detection loss,
\begin{equation}
\footnotesize
\mathcal{L}^{(k)}_{detc}(\theta;y_p)=-\sum_{i=1}^{N}[y_p^{k(i)}\log(p(y_p^{k(i)})) \\
+(1-y_p^{k(i)})\log(1-p(y_p^{k(i)}))],
\label{prop}
\end{equation}
where  $y_p^{k(i)}$  and $p({y_p}^{k(i)})$ denote the ground truth (either 1 or 0) and the probability of the $i$-th sample as the $k$-th category expression,  respectively.  $N$ is the total number of training samples. $\mathcal{L}^{(k)}_{detc}$  allows the network to generate expression-specific features for every expression. Based on the loss occurring in each sub-proposals,  the loss of the proposal module consisted of $K$ sub-branches is defined as the average
of $K$ detection losses:
\begin{equation}
\mathcal{L}_{prop}(\theta;y_p) = {\frac {1}{K}}\sum_{k=1}^K\mathcal{L}_{detc}^{(k)}.
\label{prop1}
\end{equation}
By the restraint of the proposal loss of $\mathcal{L}_{prop}$ and the attention mechanism, the proposal module obtains salient and discriminative expression-specific feature for micro-expression recognition.

\subsection{Fused expression-refined features for classification}


Generally, a simple way is to concatenate the expression-specific features into one feature vector, which is directly fed into Fully Connected Layers. However, this method will cause the high dimension and contain more trainable parameters. Motivated by the feature fusion method in \cite{Wu2019}, this work utilizes a simple but efficient method, namely, the element-wise sum as a fusion function. The efficiency evaluation about element-wise sum for fusion can be referred to Section~\ref{sec:modelablation}. Consequently, the aggregated feature representation $z'$ is defined as follows:
 \begin{equation}
z'= \sum_{k=1}^{K} z_k^*.
\label{softmax}
\end{equation}

Then, the expression-refined feature $z'$ is fed into the final classification module which consists of two fully connected layers. To avoid over-fitting, the first fully connected layer was followed by a Dropout layer (with dropout probability being 0.5). Lastly, the output of the last fully connected layer $F_i$  is activated by a $softmax$ unit. Hence, the classification loss $L_{cls}$ is expressed as:
\begin{equation}
\mathcal{L}_{cls}(\phi;y_c)=-\sum_{i=1}^{N}[y_c^i\log(\frac{exp(F_i)}{\sum_i {exp(F_i)}})],  \\
\label{cls}
\end{equation}
where $y_c^i$ is the class label for the $i$-th training instance.

By joining the proposal loss of  $\mathcal{L}_{prop}$ in Eq. (\ref{prop1}) and classification loss of $L_{cls}$ in Eq. (\ref{cls}),  a novel FR loss is proposed as follows:
\begin{equation}
\mathcal{L}(\theta, \phi;y_p,  y_c)= \lambda\mathcal{L}_{prop}(\theta;y_p)
+ \mathcal{L}_{cls}(\phi;y_c),
\label{ldisent}
\end{equation}
where the hyper-parameter $\lambda$ balances the contribution of  expression proposal and category classification.

\section{Experiments}
\subsection{Datasets}

Experiments are conducted on three commonly used spontaneous micro-expression databases: SMIC~\cite{Li2013},  CASME II \cite{Yan2014}, and SAMM~\cite{Davison2018}.

\textbf{SMIC~\cite{Li2013}:} The SMIC database contains SMIC-HS (recorded by a high-speed camera of 100 \textit{fps}),  SMIC-VIS (by a normal visual camera of 25 \textit{fps}), and SMIC-NIR (by a near-infrared camera). SMIC-HS has 164 micro-expression clips from 16 subjects,  while SMIC-VIS/SMIC-NIR consists of 71 samples from 8 participants.  Additionally, these samples in three sub-databases are annotated as \textit{Negative},  \textit{Positive}, and \textit{Surprise}. The sample resolution is $640 \times 480$ \textit{pixels} and  the facial area is around $190 \times 230$ \textit{pixels}.

\textbf{CASME~II~\cite{Yan2014} :} The CASME~II contains two versions: the first one includes 247 samples of 5 micro-expression classes (\textit{Happiness},  \textit{Surprise},  \textit{Disgust},  \textit{Repression}, and \textit{Others}), and the second has 256 samples of 7 classes (\textit{Happiness},  \textit{Surprise},  \textit{Disgust},  \textit{Sadness}, \textit{Fear}, \textit{Repression}, and \textit{Others}). All the  samples are gathered from 26 subjects. It was recorded by a camera with  200 \textit{fps}. The resolution of the samples are $640\times 480$ \textit{pixels} and the resolution of facial area is around $280 \times 340$ \textit{pixels}.

\textbf{SAMM~\cite{Davison2018}:} The SAMM database contains 159 micro-expression instances from 32 participants at 200 \textit{fps}.  The resolution of the samples are $2040 \times 1088$ \textit{pixels} and the resolution of facial area is around $400 \times 400$ \textit{pixels}. Samples in SAMM
are categorized into \textit{Happiness},  \textit{Surprise},  \textit{Disgust},  \textit{Repression},  \textit{Angry},  \textit{Fear}, \textit{Contempt}, and \textit{Others}.

\subsection{Experiment settings}

Since the apex frame annotation in SMIC database is not available, interframe-Diff method as described in Section \uppercase\expandafter{\romannumeral3} is applied to spot the apex frame. For the CASME~II and SAMM databases, the ground truth of the apex frames is directly used. The Libfacedetection~\cite{Yu2016} is utilized to crop the facial area out of onset and apex frames. TV-L1 optical flow is extracted from the onset and apex frames. Two components of optical flow images are resized to $28 \times 28$ \textit{pixels} before feeding to the Inception network. All the experiments are conducted with Ubuntu 16.04,  Python 3.6.2 with Keras 2.2.4 and Tensorflow 1.11.0 on 1 NVIDIA GTX Titan X GPU (12 GB).

\textbf{Setup:} To evaluate the effect of each module of FR, an ablation study is first designed to investigate the backbone selection, strategy selection, and fusion module selection. Second, three groups of experiments, namely, CDE experiment, CDMER experiment, and the single database experiment, are designed to validate the effectiveness of the proposed method. For fair comparison, all the experiments on CDE and the single database evaluation are conducted with Leave-One-Subject-Out (LOSO) cross-validation, where samples from one subject are held out as the testing set while all remaining samples for training. For CDMER, the model is trained on the source dataset and tested on the target dataset.

\textbf{The detail settings for model ablation experiment, CDE experiment, CDMER experiment, and the single database experiment can be referred to Section~\ref{sec:appendix}.}

\textbf{Performance metric:} Here, this work applies different performance metric for the three groups.
\begin{itemize}
    \item For CDE protocol, according to the MEGC 2019, Unweighted F1-score (\textbf{UF1}) and Unweighted Average Recall (\textbf{UAR}) are used to measure the performance of various methods on composite and individual databases.
    \item For CDMER protocol, according to~\cite{Zong2017}, Accuracy (\textbf{Acc}) and UF1 are reported for evaluating the performance.
    \item For the evaluation on the single database, Acc is used.
\end{itemize}

All results of each type of experiments are the average of at least ten rounds. \textbf{The evaluation metrics can be also referred to Section~\ref{sec:appendix}.}

\begin{table*}
\centering
\footnotesize
\caption{Ablation models. The best result is in bold.}
\label{ablation}
\vspace{0.2cm}
  \subtable[{\color{black}Backbone selection for the expression-shared feature learning.}]{
\begin{tabular}{cccc}
\hline
Model &UF1 & UAR\\
\hline
Dual-Inception \cite{Zhou2019}&0.7322& 0.7278\\
Basic Inception & \textbf{0.7360} & \textbf{0.7391} \\
\hline
  \end{tabular}
       \label{ablation:apex frame}
}
\hspace{1.2cm}
\subtable[Strategy selection for  the expression-specific feature learning.]{
\centering
\begin{tabular}{cccc}
\hline
Model &UF1 & UAR\\
\hline
FR-fc &0.7377&0.7443 \\
FR& \textbf{0.7838}&\textbf{0.7832} \\
\hline
  \end{tabular}
       \label{ablation:specific feature learning}
}
\hspace{1.2cm}
\subtable[Fusion mode selection to obtain the expression-refined feature.]{
\centering
\begin{tabular}{cccc}
\hline
Model &UF1 & UAR\\
\hline
FR-concatenated &0.7632&0.7727 \\
FR&\textbf{0.7838}&\textbf{0.7832}\\
\hline
  \end{tabular}
       \label{ablation:fusion mode}
}
\end{table*}

\subsection{Model ablation}
\label{sec:modelablation}

The ablation study is performed on the composite database of CDE protocol. Table \ref{ablation} reports the results in terms of UAR and UF1 on different models.

\begin{table*}
\caption{Performance  comparison among the handcrafted feature methods,  classical CNN methods,  state of the art  and our proposed methods on  the composite database and the individual databases on CDE protocol. In the CDE protocol,  except samples of SMIC-HS, samples in CASME II and SAMM are regrouped into three classes,~\ie~\textbf{\textit{Negative}, \textit{Positive}, and \textit{Surprise}}. Then, regrouped samples of three databases are combined into one dataset and  using LOSO validation methods to evaluate the performance. The best results are in bold.}
\label{cde_result}
\footnotesize
\begin{center}
\setlength{\tabcolsep}{2.7mm}{
\begin{tabular}{clcccccccc}

\hline
 \multirow{2}*{Groups}&\multirow{2}*{Approaches} & \multicolumn{2}{c}{Composite} &
 \multicolumn{2}{c}{SMIC-HS} & \multicolumn{2}{c}{CASME II} & \multicolumn{2}{c}{SAMM}\\
 \cline{3-10}
 && UF1&UAR&UF1&UAR&UF1&UAR&UF1&UAR\\
 \hline
\multirow{2}*{Handcrafted features}
&LBP-TOP \cite{Zhao2007}&0.5882& 0.5785 &0.2000&0.5280&0.7026&0.7429 &0.3954&0.4102\\
\cline{2-10}
&Bi-WOOF \cite{Liong2018a}&0.6296&0.6227&0.5727&0.5829&0.7805&0.8026&0.5211&0.5139\\
\hline
\multirow{9}*{Deep learning features}
&AlexNet \cite{Krizhevsky2012}&0.6933&0.7154&0.6201&0.6373&0.7994&0.8312&0.6104&0.6642\\
\cline{2-10}
&GoogLeNet \cite{Szegedy2015}&0.5573&0.6049&0.5123&0.5511&0.5989&0.6414&0.5124&0.5992\\
\cline{2-10}
&VGG16 \cite{Simonyan2014}&0.6425&0.6516&0.5800&0.5964&0.8166&0.8202&0.4870&0.4793\\
\cline{2-10}
&CapsuleNet \cite{Quang2019}&0.6520& 0.6506& 0.5820& 0.5877& 0.7068 &0.7018 &0.6209& 0.5989\\
\cline{2-10}
&OFF-ApexNet \cite{Gan2019}&0.7196&0.7096&0.6817&0.6695&0.8764&0.8681&0.5409&0.5392\\
\cline{2-10}
&Dual-Inception \cite{Zhou2019}&0.7322&0.7278&0.6645&0.6726&0.8621&0.8560&0.5868&0.5663\\
\cline{2-10}
& STSTNet \cite{Liong2019}&0.7353&0.7605&0.6801&0.7013&0.8382&0.8686&0.6588&0.6810\\
\cline{2-10}
&EMR \cite{Liu2019}&\textbf{0.7885}&0.7824&\textbf{0.7461}&\textbf{0.7530}&0.8293&0.8209&\textbf{0.7754}&0.7152\\
\cline{2-10}
&\textbf{FR (Ours)}&0.7838&\textbf{0.7832}&0.7011&0.7083&	\textbf{0.8915}&\textbf{0.8873}&0.7372&\textbf{0.7155}\\
\hline
\end{tabular}}
\end{center}
\end{table*}

\textbf{(1) Backbone selection for the expression-shared feature learning}

To better capture the subtle motion of micro-expression for our proposed model, we make a backbone selection for the expression-shared feature learning.

\textit{\textbf{a. Dual-Inception \cite{Zhou2019}:}} The model \textcolor{black}{straightforwardly} uses the mid-position frame of each sample as the apex frame for MER.

\textit{\textbf{b. Basic Inception:}} Different from~\cite{Zhou2019}, Basic Inception uses interframe difference algorithm described in Section \uppercase\expandafter{\romannumeral3} to approximately locate the apex frames for SMIC-HS database, and uses ground truth of apex frames in CASME II and SAMM.

Table \ref{ablation:apex frame} compares basic Inception to Dual-Inception on the composite database. Basic Inception outperforms Dual-Inception in terms of UAR and UF1. Thus, the Basic Inception is chosen as the backbone of our framework to learn the expression-shared feature.

\textbf{(2) Strategy selection for the expression-specific feature learning}

In the proposal module, two followed strategies are presented to learn  expression-specific features:

\textit{\textbf{a. FR-fc:}} FR-fc only uses fully-connected layers in the proposal module for the expression-specific feature learning,  and then uses the element-wise sum mode to aggregate expression-specific features of each expression category for classification.

\textit{\textbf{b. FR:}} FR differs from FR-fc using attention strategy to learn expression-specific features.

From Table \ref{ablation:specific feature learning}, FR with attention mechanism boosts the performance from 0.7377 to 0.7838 in terms of UF1, from 0.7443 to 0.7832 in terms of UAR. This suggests that attention factors in the proposal module have the capability to highlight specific characteristics and generate salient features.

\textbf{(3) Fusion mode selection to obtain the expression-refined feature}

 To validate the fusion mode for fusing the expression-specific features, this part compares element-wise sum mode used in this paper to concatenated mode.

\textit{\textbf{a. FR-concatenated:} }FR-concatenated model directly concatenates the expression-specific features.

\textit{\textbf{b. FR:}} FR model uses element-wise sum mode to obtain expression-refined features.

According to Table \ref{ablation:fusion mode}, FR outperforms the FR-concatenated. It may attribute to that element-wise sum mode alleviates the over-fitting problem as element-wise sum model more suppress the dimension of expression-refined feature than concatenated method. Consequently, the element-wise sum mode is chosen for expression-specific features fusion to obtain expression-refined features.

\subsection{Performance evaluation on CDE protocol}

Table~\ref{cde_result} compares our method to several state-of-the-art methods on CDE protocol.
These methods contain two handcrafted features (LBP-TOP \cite{Zhao2007} and Bi-WOOF \cite{Liong2018a}), six deep learning features without data augmentation technology (AlexNet \cite{Krizhevsky2012},  GoogLeNet \cite{Szegedy2015},  VGG16 \cite{Simonyan2014},  OFF-ApexNet \cite{Gan2019}, Dual-Inception \cite{Zhou2019}, and STSTNet \cite{Liong2019}), and two deep learning features with data augmentation (CapsuleNet \cite{Quang2019} and EMR \cite{Liu2019}). Specifically,  AlexNet, GoogLeNet, and VGG16 were reproduced by Liong~\etal~\cite{Liong2019} instead by the inputs of optical flow features.

\subsubsection{Comparison with handcrafted features}

When comparing to LBP-TOP, our proposed FR improves the baseline consistently with gains of 19.56\%, 50.11\%, 18.89\%, and 34.18\% in terms of UF1 for composite, SMIC-HS, CASME II, and SAMM databases,  respectively. It increases the performance of the baseline by 20.47\%,  18.03\%,  14.44\%, and 30.53\% in terms of UAR for composite,  SMIC-HS,  CASME II, and SAMM databases, respectively. FR consistently improves the Bi-WOOF by a large margin. It indirectly suggests that FR learns discriminative and meaningful features on micro-expression database, which is better than the traditional handcrafed features.

\subsubsection{Comparison with deep learning features}

Table \ref{cde_result} indicates that our proposed FR outperforms most state-of-the-art algorithms on all the databases. It is explained by that the scarcity of data causes the existing deep-depth networks the over-fitting.  The promising results further suggest that the shallow neural networks with fewer parameters alleviate the over-fitting problem in MER.
\begin{itemize}
    \item FR achieves better performance than OFF-ApexNet, CapsuleNet, Dual-Inception, and STSTNet. The similarity between FR and these three models is that they all learn features by feeding the extracted optical flows into designed shallow networks. But the major difference is that our proposed FR distills more meaningful characteristic from the expression-shared features for MER by expression-specific feature learning and fusion, while they focus on the expression-shared feature learning.

    \item As we know,  STSTNet is an extension of OFF-ApexNet, which learns deep learning features for MER with three pre-extracted optical flows. The considerable result of STSTNet indirectly suggests that the more pre-processing features input, the better performance the method obtains. Although our proposed FR and STSTNet belongs to multi-stream feature learning approach, with expression-specific feature learning and fusion, FR gains improvements of 5.03\% and 2.07\% in terms of average UF1 and UAR across four databases, respectively, though the performance of STSTNet are relatively high. The improvement indicates that exploring features with more salient and discriminative characteristic based on fewer pre-processing is a more promising approach.

    \item EMR used Eulerian Video Magnification (EVM) for magnifying micro-expression, in which EVM has been shown its effectiveness to micro-expression recognition~\cite{Li2018a}. Comparing with EMR, FR slightly degrades the performance by 0.64\% in terms of average UF1 across four databases, while increases the performance by 0.58\% in terms of average UAR across four databases. Thus, FR still obtains a competitive performance to EMR. On the other hand, FR is built on more simple pre-process~\eg~facial cropped and apex frame chosen than EMR~\eg~macro-expression samples relabeling to three categories and micro-expressions magnification.
\end{itemize}


\begin{table*}
\caption{Experimental results (UF1 / Acc) of different micro-expression features for \textbf{Type-I} of CDMER tasks. The results of feature descriptors in the benchmark are
directly extracted from \cite{Zong2019}. The best results in each experiment are in bold.}
\label{cdmer result_task1}
\begin{center}
\scriptsize
\setlength{\tabcolsep}{1.0mm}{
\begin{tabular}{clccccccc}
\hline
Groups&Feature Descriptors & \textit{Exp}.1: H$\rightarrow$V &\textit{Exp}.2: V$\rightarrow$H&\textit{Exp}.3: H$\rightarrow$N
& \textit{Exp}.4: N$\rightarrow$H & \textit{Exp}.5: V$\rightarrow$N & \textit{Exp}.6: N$\rightarrow$V & Average \\
\hline
\multirow{12}*{Handcrafted features}
&LBP-TOP\textit{(R3P8)} \cite{Zhao2007} &0.8002/0.8028& 0.5421/0.5427& 0.5455/0.5352 & 0.4878/0.5488 & 0.6186/ 0.6338 &0.6078/0.6338 &0.6003/0.6162\\
\cline{2-9}
&LBP-TOP\textit{(R1P4)} \cite{Zhao2007}&0.7185/0.7183& 0.3366/0.4024& 0.4969/0.4930& 0.3457/0.4024& 0.5480/0.5775 & 0.5085/0.5915 &0.4924/0.5332\\
\cline{2-9}
&LBP-TOP\textit{(R1P8)} \cite{Zhao2007} & 0.8561/0.8592& 0.5329/0.5366&  0.5164/0.5775&  0.3246/0.3537&  0.5124/0.5775&  0.4481/0.5070 & 0.5318/0.5686\\
\cline{2-9}
&LBP-TOP\textit{(R3P4)} \cite{Zhao2007}&0.4656/0.4930 &0.4122/0.4512& 0.3682/0.4085& 0.3396/0.4085& 0.5069/0.5915 &0.5144/0.6056 &0.4345/0.4931\\
\cline{2-9}
&LBP-SIP\textit{(R1)} \cite{Wang2014} & 0.6290/0.6338& 0.3447/0.4085& 0.3249/0.3380 &0.3490/0.4207 &0.5477/0.6056& 0.5509/0.6056 &0.4577/0.5020\\
\cline{2-9}
&LBP-SIP\textit{(R3)} \cite{Wang2014}& 0.8574/0.8592& 0.4886/0.5000& 0.4977/0.5493& 0.4038/0.4268& 0.5444/0.5915& 0.3994/0.4648& 0.5319/0.5653\\
\cline{2-9}
&LPQ-TOP\textit{(decorr=0.1)} \cite{Paeivaerinta2011} &\textbf{0.9455}/\textbf{0.9437}& 0.5523/0.5488& 0.5456/0.6197 &0.4729/0.4756 &0.5416/0.5775 &0.6365/0.6620 &0.6157/0.6379\\
\cline{2-9}
&LPQ-TOP\textit{(decorr=0)} \cite{Paeivaerinta2011}& 0.7711/0.7746 &0.4726/0.4878& 0.6771/0.6761& 0.4701/0.4817& 0.7076/0.7183& 0.6963/0.7042& 0.6325/0.6405\\
\cline{2-9}
&HOG-TOP\textit{(p=4)} \cite{Li2018a}&0.7068/0.7183 &0.5649/0.5732& \textbf{0.6977}/\textbf{0.7042} &0.2830/0.2927& 0.4569/0.4930& 0.3218/0.3662 & 0.4554/0.4847\\
\cline{2-9}
&HOG-TOP\textit{(p=8)} \cite{Li2018a}& 0.7364/0.7465 &0.5526/0.5610 &0.3990/0.4648& 0.2941/0.3232& 0.4137/0.4648 &0.3245/0.3803 &0.4453/0.4901\\
\cline{2-9}
&HIGO-TOP\textit{(p=4)} \cite{Li2018a}& 0.7933/0.8028& 0.4775/0.5061& 0.4023/0.4789 &0.3445/0.3598& 0.5000/0.5352& 0.3747/0.4085 &0.4821/0.5152\\
\cline{2-9}
&HIGO-TOP\textit{(p=8)} \cite{Li2018a}& 0.8445/0.8451 &0.5186/0.5366& 0.4793/0.5493& 0.4322/0.4390& 0.5054/0.5493& 0.4056/0.4648 &0.5309/0.5640\\
\hline
\multirow{4}*{Deep learning features}
&C3D-FC1 \textit{(Sports1M)} \cite{Tran2015}& 0.1577/0.3099 &0.2188/0.2378 &0.1667/0.3099& 0.3119/ 0.3415 &0.3802/0.4930& 0.3032/0.3662 &0.2564/0.3431\\
\cline{2-9}
&C3D-FC2 \textit{(Sports1M)} \cite{Tran2015}&0.2555/0.3662 &0.2974/0.2927 &0.2804/0.3380 &0.3239/0.3659 &0.4518/0.4789& 0.3620/0.3803& 0.3285/0.3703\\
\cline{2-9}
&C3D-FC1 \textit{(UCF101)} \cite{Tran2015} &0.3803/0.4648 &0.3134/0.3476& 0.3697/0.4789& 0.3440/0.3476 &0.3916/0.4789 &0.2433/0.2958 &0.3404/0.4023\\
\cline{2-9}
&C3D-FC2 \textit{(UCF101)} \cite{Tran2015} &0.4162/0.4648& 0.2842/0.3232& 0.3053/0.4225 &0.2531/0.2805& 0.3937/0.4789& 0.2489/0.3239 &0.3169/0.3823\\
\cline{2-9}
&\textbf{FR (Ours)}&0.7065/0.7149 &\textbf{0.5971}/\textbf{0.5968}&0.5335/0.5673& \textbf{0.5137}/\textbf{0.5200}& \textbf{0.7934}/\textbf{0.7910}&
\textbf{0.7921}/\textbf{0.7921}& \textbf{0.6561}/\textbf{0.6636}\\
\hline
\end{tabular}
}
\end{center}
\end{table*}
\begin{table*}
\caption{Experimental results (UF1 / Acc) of different micro-expression features for \textbf{Type-II} of CDMER tasks. Those results of feature descriptors in the benchmark are directly cited from \cite{Zong2019}. The best results in each experiment are in bold.}
\label{cdmer result_task2}
\begin{center}
\scriptsize
\setlength{\tabcolsep}{1mm}{
\begin{tabular}{clccccccc}
\hline
Groups&Feature Descriptors & \textit{Exp}.7: C$\rightarrow$ H &\textit{Exp}.8: H$\rightarrow$C&\textit{Exp}.9: C$\rightarrow$V
& \textit{Exp}.10: V$\rightarrow$C & \textit{Exp}.11: C$\rightarrow$N & \textit{Exp}.12: N$\rightarrow$C & Average \\
\hline
\multirow{12}*{Handcrafted features}
&LBP-TOP\textit{(R3P8)} \cite{Zhao2007}&0.3697/0.4512 &0.3245/0.4846 &0.4701/0.5070&0.5367/0.5308&0.5295/0.5211&0.2368/0.2385&0.4112/0.4555\\
\cline{2-9}
&LBP-TOP\textit{(R1P4)} \cite{Zhao2007} &0.3358/0.4451&0.3260/0.4769&0.2111/0.3521&0.1902/0.2692&0.3810/0.4366&0.2492/0.2692&0.2823/0.3749\\
\cline{2-9}
&LBP-TOP\textit{(R1P8)} \cite{Zhao2007}&0.3680/0.4390&0.3339/0.5462&0.4624/0.4930&0.5880/0.5769&0.3000/0.3380&0.1927/0.2308&0.3742/0.4373\\
\cline{2-9}
&LBP-TOP\textit{(R3P4)} \cite{Zhao2007}&0.3117/0.4390&0.3436/0.4462&0.2723/0.3944&0.2356/0.2846&0.3818/0.4.30&0.2332/0.2538&0.2964/0.3852\\
\cline{2-9}
&LBP-SIP\textit{(R1) }\cite{Wang2014}&0.3580/0.4512&0.3039/0.4462 &0.2537/0.3803&0.1991/0.2692&0.3610/0.4648&0.2194/0.2692&0.2825/0.3802\\
\cline{2-9}
&LBP-SIP\textit{(R3)} \cite{Wang2014}&0.3772/0.4268&0.3742/\textbf{0.5615}&\textbf{0.5846}/0.5915 &0.6065/0.6000&0.3469/0.3521&0.2790/0.2769&0.4279/0.4681\\
\cline{2-9}
&LPQ-TOP\textit{(decorr=0.1)} \cite{Paeivaerinta2011}&0.3060/0.4207&0.3852/0.4846&0.2525/0.3380&0.4866/0.4769&0.3020/0.3521&0.2094/0.2385& 0.3236/0.3851\\
\cline{2-9}
&LPQ-TOP\textit{(decorr=0)} \cite{Paeivaerinta2011}& 0.2368/0.4390& 0.2890/0.5154& 0.2531/0.3803 &0.3947/0.4077& 0.2369/0.3521& 0.4008/0.4154 &0.3019/0.4183\\
\cline{2-9}
&HOG-TOP\textit{(p=4)} \cite{Li2018a}& 0.3156/0.3476& 0.3502/0.4769& 0.3266/0.3521 &0.4658/0.4692& 0.3219/0.3521 &0.2163/0.2746& 0.3327/0.3791\\
\cline{2-9}
&HOG-TOP\textit{(p=8)} \cite{Li2018a}& 0.3992/0.4390& 0.4154/0.5231& 0.4403/0.4507 &0.4678/0.4769&0.4107/0.4085 &0.1390/0.2077 &0.3787/0.4177\\
\cline{2-9}
&HIGO-TOP\textit{(p=4) }\cite{Li2018a}& 0.2945/0.3963 &0.3420/0.5385 &0.3236/0.4085& 0.5590/0.5538 &0.2887/0.2958& 0.2668/0.3154 &0.3458/0.4181\\
\cline{2-9}
&HIGO-TOP\textit{(p=8)} \cite{Li2018a}&0.2978/0.4146& 0.3609/0.5000& 0.3679/0.4366& 0.5699/0.5462&0.3395/0.3380 &0.1743/0.2231& 0.3517/0.4098\\
\hline
\multirow{4}*{Deep learning features}
&C3D-FC1 \textit{(Sports1M) }\cite{Tran2015}&0.1994/0.4268& 0.2394/\textbf{0.5615}& 0.1631/0.3239& 0.1075/0.1923& 0.1631/0.3239& 0.2397/0.5615 &0.1854/0.3983\\
\cline{2-9}
&C3D-FC2 \textit{(Sports1M)} \cite{Tran2015}&0.1994/0.4268& 0.1317/0.2462& 0.1631/0.3239 &0.1075/0.1923& 0.1631/0.3239& 0.2397/0.5615& 0.1674/0.3458\\
\cline{2-9}
&C3D-FC1 \textit{(UCF101)} \cite{Tran2015}& 0.1581/0.3110& 0.1075/0.1923& 0.1886/0.3944& 0.1075/0.1923 &0.1886/0.3944& 0.2397/0.5615 &0.1650/0.3410\\
\cline{2-9}
&C3D-FC2 \textit{(UCF101)} \cite{Tran2015}& 0.1994/0.4268 &0.1705/0.1923& 0.1631/0.3239& 0.1075/0.1923& 0.1631/0.3239 &0.1075/0.1923& 0.1414/0.2753\\
\cline{2-9}
&\textbf{FR (Ours)}&\textbf{0.4670}/\textbf{0.4905}&\textbf{0.4883}/0.5380&0.5678/\textbf{0.6101}&\textbf{0.5929}/\textbf{0.6019}&\textbf{0.4399}/\textbf{0.4823}&\textbf{0.5963}/\textbf{0.6081}&\textbf{0.5254}/\textbf{0.5552}\\

\hline
\end{tabular}
}
\end{center}
\end{table*}

\subsection{Performance evaluation on CDMER benchmark}

In this experiment, the proposed FR is further evaluated by using CDMER protocol~\cite{Zong2017}. CDMER protocol contains 12 sub-experiments from the TYPE-I and TYPE-II tasks. The detail setting of CDMER is referred to Section~\ref{sec:appendix} and Table \ref{cdmer two types}. Table \ref{cdmer result_task1} and Table \ref{cdmer result_task2} compare FR to the state-of-the-art algorithms referred by ~\cite{Zong2019}. Their parameter settings are described as follows:
\begin{itemize}
    \item For LBP-TOP \cite{Zhao2007}, the uniform pattern is used. For the neighboring radius $R$ and the number of the neighboring points $P$, experiments consider three cases: $R=1, P=4$ (R1P4), $R=1, P=8$ (R1P8), $R=3, P=4$ (R3P4), and $R=3, P=8$ (R3P8).

    \item For LBP-SIP \cite{Wang2014}, the neighboring radius $R$ is set at 1 and 3, respectively.

    \item For LPQ-TOP \cite{Paeivaerinta2011}, the size of the local window in each dimension is set as $[5, 5, 5]$, and the factor for correlation model $decorr$  is set as [0.1,  0.1] and [0,  0], respectively.

    \item The number of bins $p$ is set as 4 and 8 for HOG-TOP and HIGH-TOP~\cite{Li2018a}, respectively.

    \item For three dimensional convolutional neural network (C3D) \cite{Tran2015},  the micro-expression features are extracted from the last two fully connected layers from the pre-trained Sport-1M \cite{Karpathy2014} and UCF101 \cite{Soomro2012} model.

\end{itemize}

\begin{table*}[t]
\caption{Experimental results of different micro-expression recognition approaches on the single database in terms of Acc. CASME II (5 class) means samples of the original 5 classes in CASME II are used (\textit{Happiness}, \textit{Surprise}, \textit{Disgust}, \textit{Repression} and \textit{Others}).  CASME II (4 class) and SAMM~databases mean samples of both databases are regrouped into four classes,~\ie, \textit{Negative}, \textit{Positive}, \textit{Surprise} and \textit{Others}. The best results in each experiment are highlighted in bold. All the results reported are based on the original samples without expression magnification.}
\label{single result}
\begin{center}
\footnotesize
\setlength{\tabcolsep}{1.6mm}{
\begin{tabular}{cllccccc}
\hline
Groups &Approaches&SMIC-HS&CASME II &CASME II&SAMM&Average\\
&&&(5 classes)& (4 classes)&&\\
\hline
\multirow{22}*{Handcrafted features}
&LBP-TOP \cite{Zhao2007}&0.4878&-&0.4090$^*$&0.4150$^*$&0.4357\\
\cline{2-7}
&OSF + OS weighted LBP-TOP \cite{Liong2016}&0.5244&-&-&-&0.5244\\
\cline{2-7}
&OS \cite{Liong2014a}&0.5356&-&-&-&0.5356\\
\cline{2-7}
&OS weighted LBP-TOP \cite{Liong2014}&0.5366&0.4200&-&-&0.4783\\
\cline{2-7}
&STM \cite{Ngo2014}&0.4434&0.4378&-&-&0.4406\\
\cline{2-7}
&LBP-MOP \cite{Wang2015}&0.5061&0.4413&-&-&0.4737\\
\cline{2-7}
&LBP-TOP + ROIs \cite{Liong2018}&0.5400&0.4600&-&-&0.5000\\
\cline{2-7}
&LBP-SIP \cite{Wang2014}&0.4451&0.4656&0.4570$^*$&0.4170$^*$&0.4462\\
\cline{2-7}
&LBP-TOP + DMDSP \cite{Ngo2017}&0.5800&0.4900&-&-&0.5350\\
\cline{2-7}
&LBP-TOP + Adaptive MM \cite{Park2015}&0.5191&-&-&-&0.5191\\
\cline{2-7}
&HFOFO \cite{Happy2019}&0.5183&0.5664&-&-&0.5424\\
\cline{2-7}
&STCLQP \cite{Huang2016}&0.6402&0.5839&-&-&0.6121\\
\cline{2-7}
&STLBP-IP \cite{Huang2015}&0.5793&0.5951&0.5510$^*$&0.5680$^*$&0.5734\\
\cline{2-7}
&MMFL \cite{He2017a}&0.6315&0.5981&-&-&0.6148\\
\cline{2-7}
&Hierarchical STLBP-IP \cite{Zong2018a}&0.6078&0.6383&-&  - &0.6231\\
\cline{2-7}
&STRBP \cite{Huang2017}&0.6098&0.6437&-&-&0.6268\\
\cline{2-7}
&DiSTLBP-RIP \cite{Huang2019}&0.6341&0.6478&-&-&0.6410\\
\cline{2-7}
&MDMO \cite{Liu2016}&0.6150$^*$&-&0.5100$^*$&  - &0.5630\\
\cline{2-7}
&FDM \cite{Xu2017}&0.5488&0.4593&0.4170$^*$& - &0.4750\\
\cline{2-7}
&Bi-WOOF \cite{Liong2018a}&0.5930$^*$&-&0.5890$^*$&0.5980$^*$&0.5930\\
\cline{2-7}
&OF Maps \cite{Allaert2017}&-&\textbf{0.6535}&-&-&0.6535\\
\cline{2-7}
&Bi-WOOF + Phase \cite{Liong2017}&\textbf{0.6829}&0.6255&-&-&\textbf{0.6542}\\
\cline{2-7}
&HIGO \cite{Li2018a}&0.6524&0.5709&-&-&0.6117\\
\hline
\multirow{8}*{Deep learning features}
&Image-based CNN \cite{Takalkar2017}&0.3120$^*$&-&0.4440$^*$&0.4360$^*$&0.3973\\
\cline{2-7}
&3D-FCNN \cite{Li2018}&0.5549&0.5911&-&-&0.5730\\
\cline{2-7}
&CNN + LSTM \cite{Kim2016}&-&0.6098&-&-&0.6098\\
\cline{2-7}
&STRCN-A \cite{Xia2020}&0.4810&-&0.4710&0.4880&0.4800\\
\cline{2-7}
&STRCN-G \cite{Xia2020}&0.5760&-&0.6210&\textbf{0.6420}&0.6130\\
\cline{2-7}
&\textbf{FR (Ours)}&0.5790&0.6285&\textbf{0.6838}&0.6013&0.6232\\
\hline
\multicolumn{5}{l}{\scriptsize{$*$ means that we directly
extracted the result from \cite{Xia2020} as the original papers did not report these relevant results.}}\\
\end{tabular}
}
\end{center}
\end{table*}

\subsubsection{Comparison with handcrafted features}
As seen from Table \ref{cdmer result_task1} and Table \ref{cdmer result_task2}, our proposed FR outperforms all handcrafted features in terms of both unweighted F1-score and accuracy. Additionally, two important observations are concluded as follows:
\begin{itemize}
    \item The performance of handcrafted features fluctuates greatly with the adjustment of parameters. For example, in Table \ref{cdmer result_task1}, as the results of  LBP-TOP in \textit{Exp.1} of Type-I task showed, it gains considerable performance (0.8561 / 0.8592) under R1P8, but worse result (0.4656 / 0.4930) under R3P4. The same to LBP-SIP in \textit{Exp.9} of Table \ref{cdmer result_task2}.

    \item Varied parameters lead to perform unsteadily for Engineered features on all tasks. Instead, our proposed FR achieves the stable performance. For example, although LPQ-TOP\textit{(decorr=0.1)} obtains better results (with UF1 being 0.9455, with Acc being 0.9437)  than our proposed FR (with UF1 being 0.7065, with Acc being 0.7149) in \textit{Exp.1} of TYPE-I task, but dramatically degrades performance in \textit{Exp.3} of TYPE-II task. Therefore, the both two observations indicate that our proposed method performs more stable and robust (database-invariant) to different situations than engineered features.
\end{itemize}
\subsubsection{Comparison with deep learning features}
FR outperforms image-based C3D, which used the small scale of micro-expression data and low intensity of micro-expressions.  It suggests that exploiting available information~\eg~optical flow and shallow network will benefit to MER.

Finally, the results of experiments on CDE and CDMER protocols demonstrate that FR performs robust recognition in complex situations,~\eg~cross-database MER.

\subsection{Performance evaluation on the single database}

Table \ref{single result} compares our proposed FR to the state-of-the-art algorithms on four single micro-expression database.

\subsubsection{Comparison with handcrafted features}

FR outperforms most of the handcrafted features listed in Table \ref{single result} in terms of average recognition accuracy, except STRBP \cite{Huang2017}, DiSTLBP-RIP \cite{Huang2019}, OF Maps \cite{Allaert2017}, and Bi-WOOF with Phase \cite{Liong2017}. It suggests that although almost many of the research fields in computer vision are focusing on the deep learning century, traditional machine learning features play an importance role in MER. It is explained by the scale of data.

As we know, OF Maps and Bi-WOOF with Phase need more process to extract better features and are based on professional knowledge. Specifically, they all need to compute direction and magnitude statistical profiles of optical flow, and Bi-WOOF with Phase also needs using Riesz transform to extract phase information. On the other hand, LBP-based features of STRBP and DiSTLBP-RIP are based on the entire sequence of samples, which also need to use temporal interpolation method (TIM) \cite{Zhou2012} to normalize each video for performance improvement. In contrast, our proposed FR only needs simple pre-processing,~\eg~face detection, apex frame chosen, and optical flow extraction.

\subsubsection{Comparison with deep learning features}

Furthermore, Table \ref{single result} compares our proposed FR to Image-based CNN~\cite{Takalkar2017}, 3D-FCNN \cite{Li2018}, CNN with LSTM \cite{Kim2016}, and Spatiotemporal Recurrent Convolutional Networks (STRCN) \cite{Xia2020}, of which experiments were conducted on the same database and micro-expression categories to ours. Specifically, STRCN contains STRCN-A and STRCN-G. The first one vertorizes one channel of a frame into a column of the matrix for the appearance features, while the latter one uses optical flow images as the input to train the model. First, Table \ref{single result} shows all the handcrafted features outperform the Image-based CNN feature \cite{Takalkar2017}, because Image-based CNN feature ignored the temporal information. On the other hand, other works on CNN~\cite{Li2018,Kim2016} demonstrate temporal information significantly boosts the performance of CNN. Thus, these results suggest that when designing CNN for MER, temporal information should be considered. Furthermore, FR gains a considerable improvement of 14.32\% by comparing with STRCN-A. FR obtains competitive performance to STRCN-G. The comparison suggests that geometric-based features may become a complementary information to deep learning model. Additionally, the comparison motivates us to consider how the geometric-based features are embedded in our FR model in future. Finally, our FR suppresses all the deep learning features. The comparison results demonstrate that both expression-specific feature and feature fusion contribute the discriminative information to MER in deep learning methods.

\begin{table}[t!]
 \centering
\caption{Performance comparison of the composite database among Basic Inception and FR on CDE protocol.}

\setlength{\tabcolsep}{3.0mm}{
\begin{tabular}{ccc}
\hline
Model  &UF1 & UAR\\
\hline
Basic Inception & 0.7360 & 0.7391\\
\textbf{FR}&\textbf{0.7838}&\textbf{0.7832}\\
\hline
\end{tabular}}
\label{result_cmp_on_Basic and FR}
\end{table}

 \begin{figure}[t!]
 \centering
   \subfigure[The confusion matrix of composite database in Basic Inception.]
    {
        \includegraphics[width=0.20\textwidth]{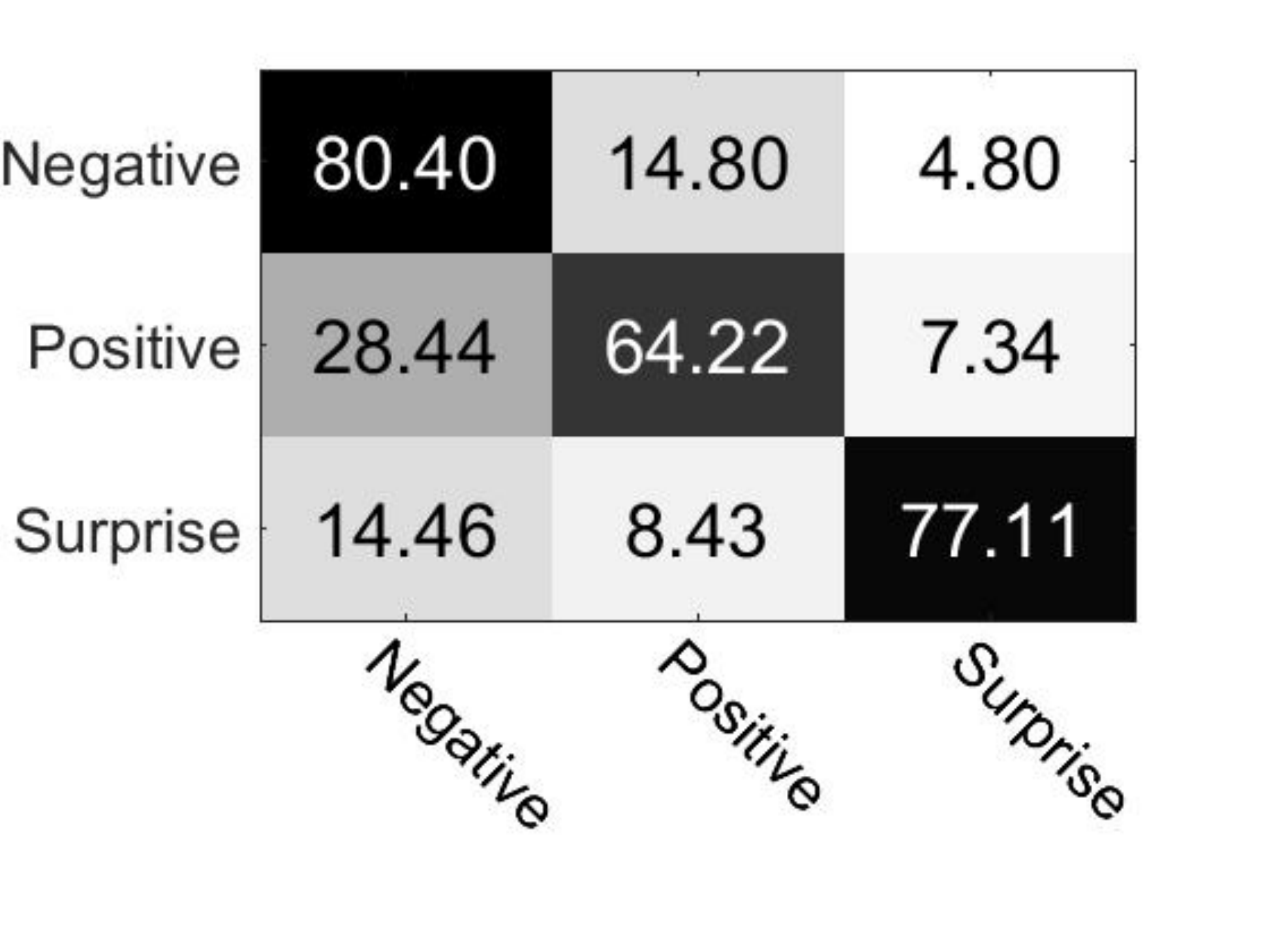}
        \label{cm_basic}
    }
   \subfigure[The confusion matrix of composite database in FR.]
    {
        \includegraphics[width=0.20\textwidth]{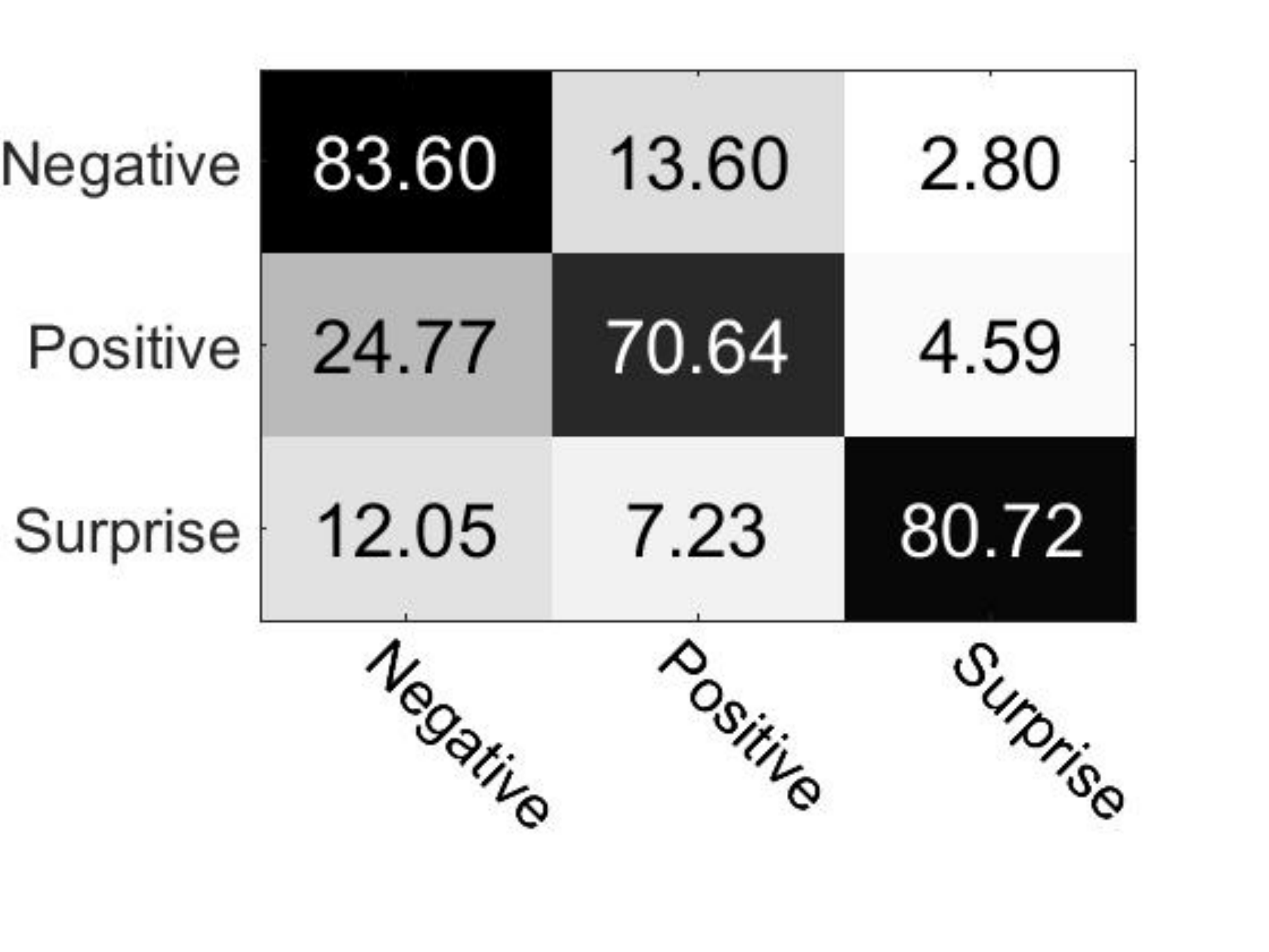}
        \label{cm_fr}
    }
 \caption{The confusion matrices of composite database in Basic Inception and FR.}
  \label{cm}
\end{figure}

 \begin{figure*}[t!]
 \centering
   \subfigure[Expression-shared feature distribution of Basic Inception.]
    {
        \includegraphics[width=0.45\textwidth]{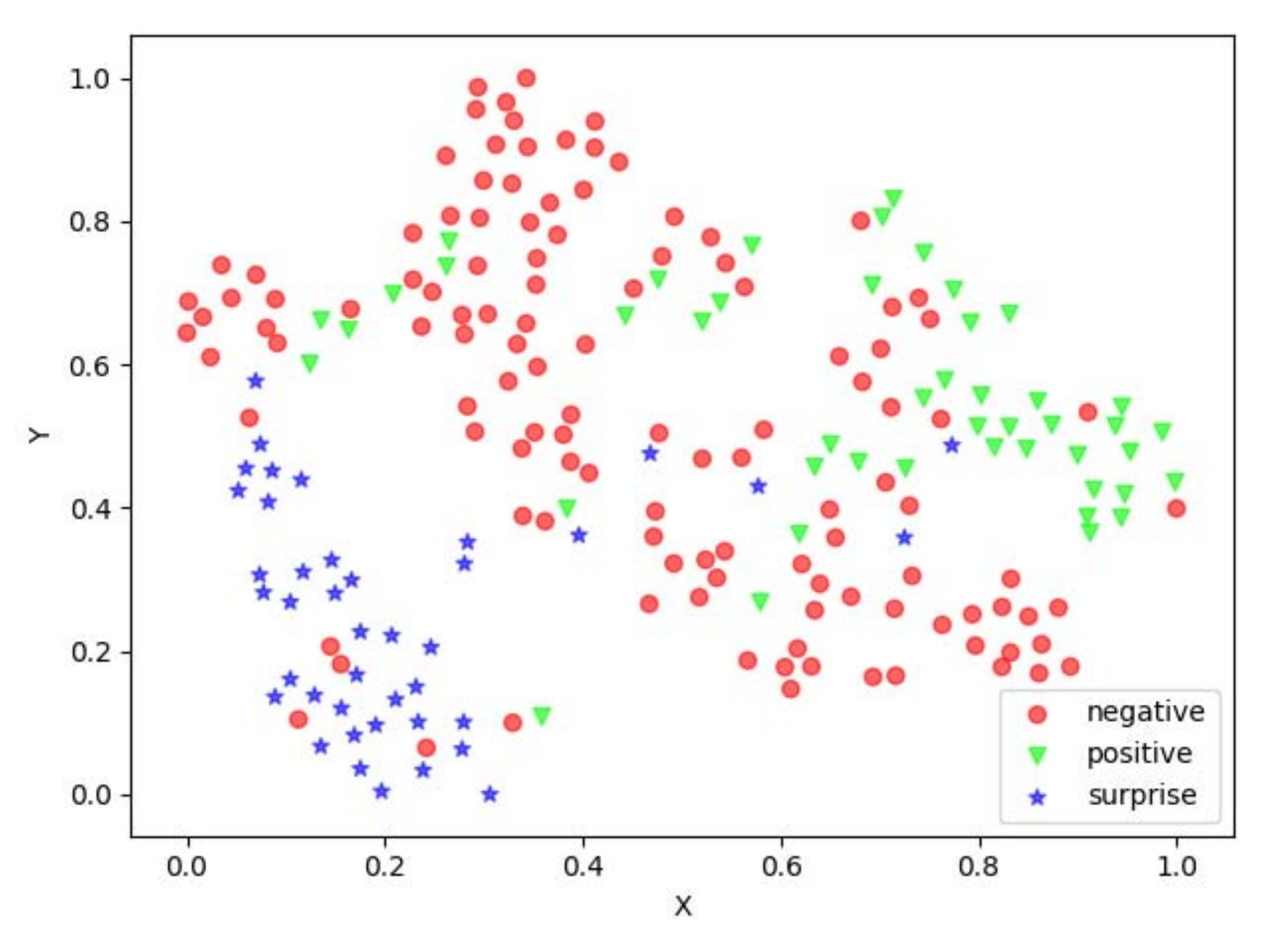}
        \label{Basic_feature_distribution}
    }
  \hspace{1cm}
   \subfigure[Expression-refined feature distribution of FR.]
    {
        \includegraphics[width=0.45\textwidth]{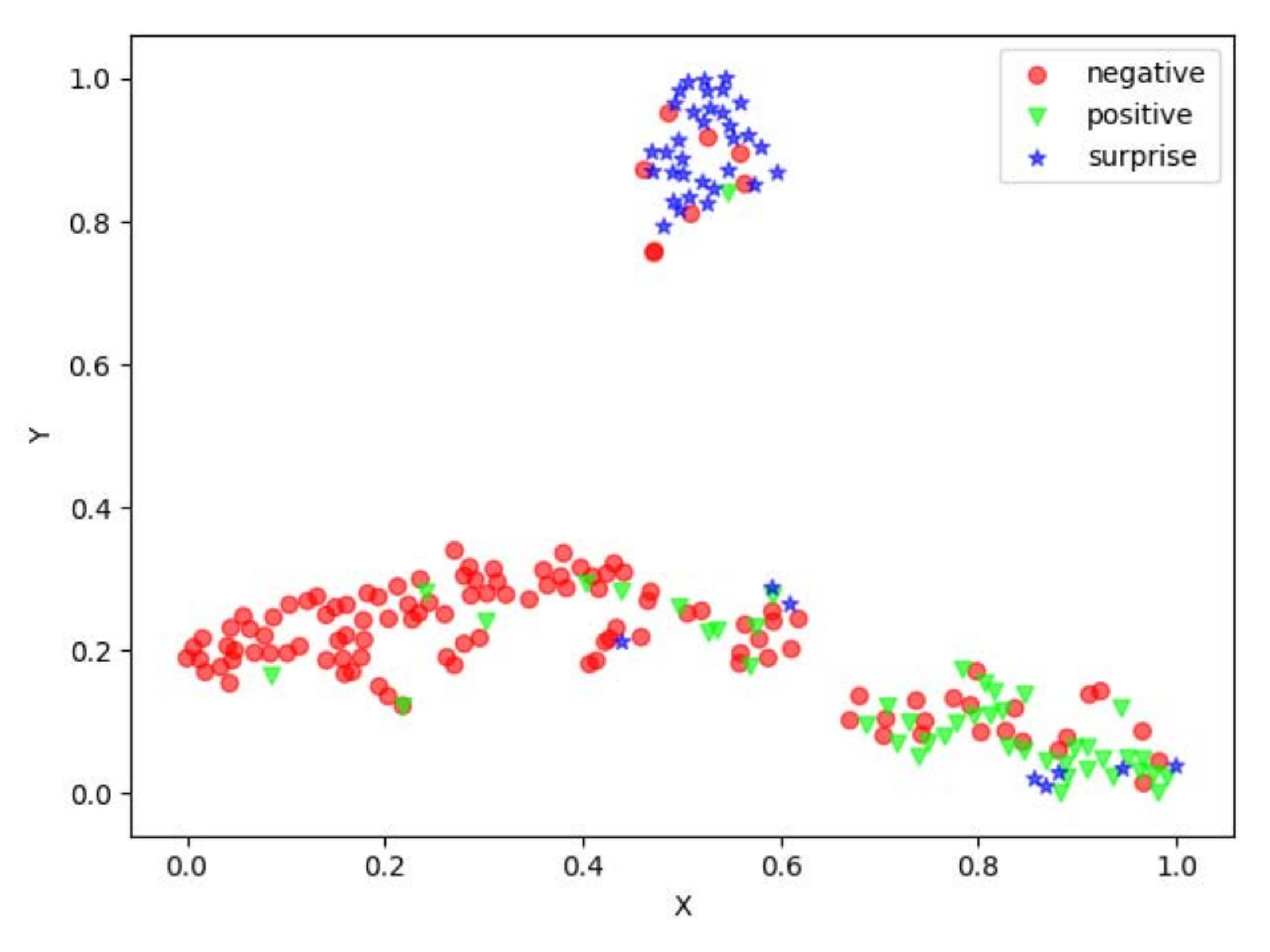}
        \label{FR_feature_distribution}
    }
 \caption{Feature distributions of Basic Inception and FR.}
  \label{feature_distribution}
\end{figure*}

\subsection{Analysis on feature's salience and discrimination}

As previously described in Section~\ref{sec:ProposedMethod}, expression-specific feature learning and fusion aim to learn the salient and discriminative feature. Here, to better reveal the effect of expression-specific feature learning and fusion module to FR, the comparison is conducted to compare the expression-refined features to the expression-shared features obtained by the Basic Inception on the composite database. The CDE protocol is used.

Table \ref{result_cmp_on_Basic and FR} reports comparison results in terms of UF1 and UAR. It is seen that with expression-specific feature learning and fusion module FR obtains the significant improvement which improves the Basic Inception from 73.60\% to 78.38\% in terms of UF1, from 73.91\% to 78.32\% in terms of UAR. This suggests that expression-specific feature learning and fusion module is the most contributed module for our FR. Furthermore, Fig. \ref{cm} shows the confusion matrices of Basic Inception and FR on each micro-expression category. As seen from Fig. \ref{cm}, FR obtains the accuracy of 83.60\%, 70.64\%, and 80.72\% for negative, positive, and surprise, respectively. When adding expression-specific feature learning and fusion module to FR, FR improves Basic Inception consistently with gains of 3.2\%, 6.42\%, and 3.61\% for negative, positive, and surprise, respectively. This indicates that the expression-refined features can improve the \textbf{salience} in each class and highlight the specific characteristics of each type of expression, and thus perform better than the expression-shared features in individual categories.



In order to get more reasonable and stable feature distribution visualization, 34 subjects are randomly chosen, where 282 samples are used for training and the rest for testing. Fig. \ref{feature_distribution} shows the feature distribution of Basic Inception and FR from the testing samples, where all features are mapped to 2D using t-SNE \cite{LaurensvanderMaaten2008}. For the Basic Inception model, the features are from the concentrated layer before the classifier, while for FR, the expression-refined features are obtained before the finally classification module. It is observed that the feature representations learned by FR are better separated, making the intra-class distribution more compact and the inter-class distribution more dispersed. These visualization results indicate that expression-refined features learned by FR are more \textbf{discriminative} than the expression-shared features learned by Basic Inception.



\begin{table*}
\caption{The Accuracy (Acc), learnable Parameters (Parameter), and Execution Time (ET)  comparison among backbone of Basic Inception and FR model, where learnable parameters include the weights and biases of the network, and Execution Time means the average training and testing times on single database evaluation.}
\label{compexityofspecific}
\begin{center}
\footnotesize
\setlength{\tabcolsep}{3mm}{
\begin{tabular}{l|ccc|ccc|ccc}
\hline
 \multirow{3}*{Methods}&\multicolumn{3}{c|}{SMIC-HS} &
 \multicolumn{3}{c|}{CASME II (4 classes)} & \multicolumn{3}{c}{SAMM}\\
 \cline{2-10}
 &Acc&Parameter
&ET&Acc&Parameter
&ET&Acc&Parameter
&ET\\
&&($Million$)&($s$)&&($Million$)&($s$)&&($Million$)&($s$)\\
 \hline
Basic Inception&0.5612&6.4803&15.0610&0.6601&6.4813&32.9442&0.5783&6.4813&18.5792\\
FR&0.5790&10.2418&15.5559&0.6838&11.4231&36.7728&0.6013&11.4231&20.6564\\
\hline
\end{tabular}}
\end{center}
\end{table*}

 \subsection{Analysis on the complexity of the expression-specific feature learning and fusion}

As previously described in Section~\ref{sec:ProposedMethod}, FR with the backbone of Basic Inception learns expression-shared feature, while FR with the expression-specific feature learning and fusion module learns and aggregates expression-specific feature. To analyze the complexity of the expression-specific feature learning and fusion of FR, Table \ref{compexityofspecific} compares FR to Basic Inception in terms of accuracy (Acc), learnable parameters (Parameters), and execution time (ET) on three micro-expression databases. 

According to Table \ref{compexityofspecific}, compared with the number of parameters in Basic Inception, the learnable parameters size indeed increases significantly in FR which is caused by the expression-specific feature learning and fusion. In other words, as SMIC-HS contains three categories, FR includes three expression-specific feature learning sub-branches. It leads to more 3.7615 $million$ parameters to learn, but only more 0.4949 $s$ to execute for FR when compared with Basic Inception. It reveals that although the number of parameters increases by adding expression-specific feature learning and fusion module, the additional execution time is still acceptable. The comparison results and analysis validate expression-specific feature learning and fusion is still efficient and effective for MER.

\begin{table}[t!]
\caption{Accuracy (Acc) of SMIC-HS database under different protocols of FR. $Exp.i$ is the number of the experiment in CDMER protocol and $S$ and $T$ are the source and target micro-expression databases, respectively. \textit{C}, \textit{H}, \textit{V}, and \textit{N} denote the CASME II, SMIC-HS, SMIC-VIS, and SMIC-NIR databases, respectively. The best results are highlighted in bold.}
\label{smichs_3protocols}
\begin{center}
\footnotesize
\setlength{\tabcolsep}{2.6mm}{
\begin{tabular}{cccc}
\hline
Protocols &Experiments &Acc & Validation rule\\
\hline
CDE& -&\textbf{0.6951}&LOSO\\
\hline
\multirow{3}*{CDMER}
& \textit{Exp}.2: V$\rightarrow$H&0.5968& 5-fold\\
& \textit{Exp}.4: N$\rightarrow$H&0.5200& 5-fold\\
& \textit{Exp}.7: C$\rightarrow$H&0.4905& 5-fold\\
\hline
Single database& -&0.5790& LOSO\\
\hline
\end{tabular}}
\end{center}
\end{table}
\subsection{Discussion on three protocols}

This previous parts extensively discuss on results under CDE, CDMER, and the single database evaluation protocols. According to the previously discussed results, several observations can be concluded in the following. First, algorithm in the single database of SMIC-HS obtained worse results than by using CDMER protocol. Second, algorithm failed to achieve same or similar performance under CDE protocol and in the single database experiment. The following will discuss these two observations.


\textbf{(1) Why does algorithm under single database evaluation not always outperform under CDMER protocol?}

According to Table \ref{smichs_3protocols}, the performance of SMIC-HS database under the single database protocol is only better than that of CDMER protocol in the $Exp.4: N\rightarrow H$ and $Exp.7: C\rightarrow H$. In contrast, it works worse than CDMER in the $Exp.2: V\rightarrow H$. In $Exp.4: N\rightarrow H$ and $Exp.7: C\rightarrow H$, the data between SMIC-NIR/CASME II and SMIC-HS is heterogeneous. In other words, the data were collected under different conditions. In contrast, in $Exp.2: V\rightarrow H$, samples in SMIC-VIS recorded by a normal visual camera are more similarly to the sample in SMIC-HS database. As well, both databases contains the same participants.

The recent works~\cite{Liu2019,Peng2018,Wang2018} have indicated that the model leveraging on macro-expression can boost the deep neural network. The way may benefit to micro-expression recognition: Compared with collecting more micro-expressions of a person to reveal one's true feeling, it is much easier to collect the person's macro-expressions. It motivates us to leverage macro-expressions transferring learning method to boost our proposed FR in future. Additionally, leveraging the subject information and transfer learning strategy  allows us to obtain better recognition performance.

\textbf{(2) Why does algorithm under CDE protocol outperform that under the single database evaluation?}

This is explained by that increasing number of samples from other micro-expression databases and the optical flow feature contribute to FR. As we know, more samples can partly avoid from the overfitting problem. On the other hand,  in our framework, optical flow is fed into FR. Optical flow mainly focuses on extracting motion feature of samples, which mostly suppresses the facial identity. Consequently, besides transfer learning and other domain adaption mechanisms, adding samples from different micro-expression databases and also utilizing proper motion feature to train the model is a considerable approach to obtain better performance of MER.

\section{Conclusion}

In this paper, we propose a novel approach for micro-expression recognition, which involves three feature refinement stages: expression-shared feature learning, expression-specific feature distilling, and expression-specific feature fusion. Different from the existing deep learning methods in MER which focus on learning expression-shared features, our approach aims to learn a set of expression-refined features by expression-specific feature learning and fusion. To make the learned feature more salient and discriminative, we propose a constructive but straightforward attention strategy and a simple proposal loss in expression proposal module for expression-specific feature learning. Experiments on three publicly available micro-expression databases and three different evaluation scenarios testify the efficacy of our proposed approach.  In the future,  we will consider an end-to-end approach for MER, find more effective ways to enrich the micro-expression samples, and use transfer learning from the large-scale databases to make benefit for MER.

\section{Appendix:
experiment settings and evaluation metrics for experiments
}
\label{sec:appendix}
\subsection{Detail experiment settings for four types of experiments}
Firstly, to easily understand the settings of the four types of experiments, we give the detailed description of the experiment settings as follows.

\subsubsection{Settings of model ablation}

Ablation study is conducted on the CDE protocol in MEGC 2019 \cite{See2019}, where this study focuses on choosing the backbone for expression-shared feature learning, the strategy for expression-specific feature learning, and the fusion mode for expression-specific features aggregating.

According to the CDE protocol, LOSO validation is used to evaluate the model performance. SMIC-HS,  CASME~II, and SAMM are merged into one dataset. To make these databases share the common types of expression,  original emotion classes are re-grouped into three main categories,~\ie~\textit{Positive}, \textit{Surprise}, and \textit{Negative}. Specifically, samples of \textit{Happiness} are given \textit{Positive} labels while the labels of \textit{Surprise} samples are unchanged. Samples of \textit{Disgust},  \textit{Repression}, \textit{Anger}, \textit{Contempt}, \textit{Fear}, and \textit{Sadness} are grouped into \textit{Negative}. Table \ref{cde data} presents the detail information about three databases used for CDE protocol.

In the ablation experiment, without momentum, the batch size, learning rate, and loss function weight factor $\lambda$  are set as 32,  0.001, and 0.85,  respectively.

\begin{table}[h!]
\footnotesize
\caption{The sample distribution on three classes (\textit{Positive}, \textit{Negative}, and \textit{Surprise}) of three databases and composite database on the CDE protocol. As LOSO validation rule is used in CDE experiment, subject number is given in the table.}
\label{cde data}
\begin{center}
\setlength{\tabcolsep}{0.5mm}{
\begin{tabular}{cccccc}
\hline
 \multirow{2}*{Database} &\multicolumn{4}{c}{Micro-Expression Category} &\multirow{2}*{Subjects} \\\cline{2-5}
 &\textit{Negative} &\textit{Positive}& \textit{Surprise} & \textbf{Total} \\
\hline
SMIC-HS~\cite{Li2013} & 70 &51&43&164&16\\
\hline
CASME II~\cite{Yan2014} &88$^{\dag}$ &32&25&145&24\\
\hline
SAMM ~\cite{Davison2018}& 92$^{\S}$&26&15&133&28 \\
\hline
Composite&250&109&83&442&68\\
\hline
\multicolumn{5}{l}{\scriptsize{${\dag}$ Negative class of CASME II: Disgust and Repression.}}\\
\multicolumn{5}{l}{\scriptsize{${\S}$ Negative class of SAMM: Anger, Contempt, Disgust, Fear and Sadness.}}\\
\end{tabular}}
\end{center}
\end{table}

\begin{table}[h!]
\footnotesize
\caption{Two types of CDMER tasks. It contains 12 CDMER experiments. Each source to target experiments of CDMER is denoted by $Exp.i: S \rightarrow T$, where $Exp.i$ is the number of this experiment and $S$ and $T$ are the source and target micro-expression databases, respectively. \textit{C}, \textit{H}, \textit{V}, and \textit{N} denote the CASME II, SMIC-HS, SMIC-VIS, and SMIC-NIR databases, respectively.}
\label{cdmer two types}
\begin{center}
\setlength{\tabcolsep}{2mm}{
\begin{tabular}{cccc}
\hline
Type & CDMER Task &Source Database&Target Database\\
\hline

\multirow{6}*{Type-I}
&\textit{Exp}.1: H$\rightarrow$V &SMIC-HS&SMIC-VIS\\
&\textit{Exp}.2: V$\rightarrow$H &SMIC-VIS&SMIC-HS\\
&\textit{Exp}.3: H$\rightarrow$N &SMIC-HS&SMIC-NIR\\
&\textit{Exp}.4: N$\rightarrow$H &SMIC-NIR&SMIC-HS\\
&\textit{Exp}.5: V$\rightarrow$N &SMIC-VIS&SMIC-NIR\\
&\textit{Exp}.6: N$\rightarrow$V &SMIC-NIR&SMIC-VIS\\
\hline
\multirow{6}*{Type-II}
&\textit{Exp}.7: C$\rightarrow$H &CASME II&SMIC-HS\\
&\textit{Exp}.8: H$\rightarrow$C &SMIC-HS&CASME II\\
&\textit{Exp}.9: C$\rightarrow$V &CASME II&SMIC-VIS\\
&\textit{Exp}.10: V$\rightarrow$C &SMIC-VIS&CASME II\\
&\textit{Exp}.11: C$\rightarrow$N &CASME II&SMIC-NIR\\
&\textit{Exp}.12: N$\rightarrow$C &SMIC-NIR&CASME II\\
\hline
\end{tabular}}
\end{center}
\end{table}

\begin{table}[h!]
\footnotesize
\caption{The sample distribution on three classes (\textit{Positive}, \textit{Negative}, and \textit{Surprise}) of CASME II and SMIC databases on CDMER protocol. Five-fold cross validation rule is used in CDMER experiment.}
\label{cdmer data}
\begin{center}
\setlength{\tabcolsep}{2.5mm}{
\begin{tabular}{lccccc}
\hline
 \multirow{2}*{Database} &\multicolumn{4}{c}{Micro-Expression Category} \\\cline{2-5}
 &\textit{Negative} &\textit{Positive}& \textit{Surprise} & \textbf{Total} \\
\hline
SMIC-HS~\cite{Li2013} & 70 &51&43&164 \\
\hline
SMIC-NIR~\cite{Li2013} & 23 &28&20&71\\
\hline
SMIC-VIS~\cite{Li2013} & 23 &28&20&71\\
\hline
CASME II~\cite{Yan2014} &73$^{\dag}$ &32&25&130\\
\hline
\multicolumn{5}{l}{\scriptsize{${\dag}$ Negative class of CASME II: Disgust, Sadness and Fear.}}\\
\end{tabular}}
\end{center}
\end{table}

\subsubsection{Settings of CDE experiment}

Our proposed FR compares with the baseline of MEGC 2019 \cite{Pfister2011, Zhao2007}, three popular deep learning networks \cite{Krizhevsky2012, Szegedy2015, Simonyan2014}, and several state-of-the-art methods \cite{Liong2018a, Gan2019, Zhou2019,Liong2019, Liu2019,Quang2019}. In the experiments, we use the same expression grouped rules and parameters setting for CDE evaluation to the experiment of model ablation.

\subsubsection{Settings of CDMER experiment }

This paper evaluates the proposed FR with~\cite{Zhao2007,  Wang2014,  Paeivaerinta2011,  Li2018a, Tran2015} under CDMER protocol \footnote{http://aip.seu.edu.cn/cdmer/} \cite{Zong2017}. Five-fold cross validation.

\textbf{CDMER protocol:} Four publicly available micro-expression databases are used in the CDMER benchmark: SMIC-HS,  SMIC-VIS,  SMIC-NIR, and CASME~II. Different from CDE protocol, in each experiment of CDMER, two of four databases are chosen, where one is used as source database and another as target database. Thus, there are 12 sub-experiments in CDMER. The detail setup is depicted in Table \ref{cdmer two types}. Each source to target sub-experiment of CDMER is denoted by $Exp.i: S \rightarrow T$, where $Exp.i$ is the number of this sub-experiment, $S$ and $T$ are the source and target databases, respectively.

Table \ref{cdmer data} describes the sample distribution for CDMER experiments. The re-group rule for \textit{Positive} and \textit{Surprise} classes is the same to the CDE protocol, while the samples with \textit{Disgust}, \textit{Sadness}, and \textit{Fear} classes belong to \textit{Negative} class. In the CDMER experiment, the learning rate is set as 0.0005,  with the momentum rate of 0.8. Batch size and $\lambda$ are set as 32 and 0.85, respectively.


\begin{table*}[t!]
\footnotesize
\caption{The sample distribution on four classes (\textit{Positive}, \textit{Negative}, \textit{Surprise}, and \textit{Others}) of three databases on the single database evaluation. As LOSO validation rule is used in the single database experiment, subject number is also given in the table.}
\label{single data}
\begin{center}
\setlength{\tabcolsep}{1.0mm}{
\begin{tabular}{lcccccc}
\hline
 \multirow{2}*{Database} &\multicolumn{5}{c}{Micro-Expression Category} &\multirow{2}*{Subjects}\\\cline{2-6}
 &\textit{Negative} &\textit{Positive}& \textit{Surprise} & \textit{Others}& \textbf{Total} \\
\hline
SMIC-HS~\cite{Li2013}&70&51&43&-&164&16 \\
\hline
CASME II~\cite{Yan2014}&\multirow{2}*{73$^{\dag}$}&\multirow{2}*{32}&\multirow{2}*{25}&\multirow{2}*{126$^{\S}$}&\multirow{2}*{256}&\multirow{2}*{26}\\
(4 classes)&&&&&&\\
\hline
SAMM \cite{Davison2018}&92$^{\sharp}$&26&15&26&159&32\\
\hline
\multicolumn{6}{l}{\scriptsize{${\dag}$ Negative class of CASME II: Disgust, Sadness and Fear.}} \\
\multicolumn{6}{l}{\scriptsize{${\S}$ Others class of CASME II: Repression and Others. }}\\
\multicolumn{6}{l}{\scriptsize{${\sharp}$ Negative class of SAMM: Disgust, Anger, Contempt, Fear and Sadness. }}\\
\end{tabular}}
\end{center}
\end{table*}

\begin{table*}[t!]
\footnotesize
\caption{\textcolor{black}{The sample distribution of CASME II (5 classes).  As LOSO validation rule is used in the single database experiment, subject number is also given in the table.}}
\label{casme2_5classes}
\begin{center}
\setlength{\tabcolsep}{0.7mm}{
\begin{tabular}{lccccccc}
\hline
 \multirow{2}*{Database} &\multicolumn{6}{c}{Micro-Expression Category} &\multirow{2}*{Subjects}\\
 \cline{2-7}
 &\textit{Repression} &\textit{Happiness}& \textit{Surprise} & \textit{Disgust}& \textit{Others} &\textbf{Total} \\
\hline
CASME II~\cite{Yan2014}&\multirow{2}*{27}&\multirow{2}*{32}&\multirow{2}*{25}&\multirow{2}*{64}&\multirow{2}*{99}&\multirow{2}*{247}&\multirow{2}*{26}\\
\textcolor{black}{(5 classes)}&&&&&&\\
\hline
\end{tabular}}
\end{center}
\end{table*}

\subsubsection{Settings of the single database experiment }

Leave-one-subject-out (LOSO) protocol is used. The parameters settings of the single database experiment are the same to CDMER evaluation. Table~\ref{single data} lists the detailed information about ``CASME II (4 classes)'', SAMM and SMIC-HS databases.

For CASME II, there are two versions of the database. The first version contains 247 samples of 5 classes, while the second version involves 256 samples labeled as 7 classes. To make fair comparison, we use these two versions of CASME II for the single database experiment. For the first version, we directly use 5 micro-expression categories. For convenience, we abbreviate the first version of CASME II  with 5 classes as ``CASME II (5 classes)'' in our experiment. Table~\ref{casme2_5classes} describes the detail information of ``CASME II (5 classes)''. Following~\cite{Liu2016, Wang, Zong2018a, Xia2020}, for the second version, the samples with \textit{Happiness} class belong to \textit{Positive} class, while the samples with \textit{Surprise} class remain the original label. The samples with \textit{Disgust}, \textit{Anger}, \textit{Contempt}, \textit{Fear}, and \textit{Sadness} classes are assigned to \textit{Negative} label. Samples of \textit{Repression} are labeled to \textit{Others} class. Lastly, the samples in the second version of CASME II  are categorized into four classes. Here, we denote it as ``CASME II (4 classes)'' in the experiment. The same category protocol is used for SAMM database.

\subsection{Detail description for performance metrics}

The Acc, UAR, and UF1 metrics are defined as follows:

\begin{equation}
\footnotesize
Acc := \frac{\sum_{k=1}^{K}\sum_{m=1}^{M}TP_k^{(m)}}{\sum_{k=1}^{K}\sum_{m=1}^{M}FP_k^{(m)}+\sum_{k=1}^{K}\sum_{m=1}^{M}TP_k^{(m)}},
\label{acc}
\end{equation}

\begin{equation}
\footnotesize
UF1 = \frac{1}{K}{F1_k},
\label{Uf1}
\end{equation}
and
\begin{equation}
\footnotesize
UAR = \frac{1}{K}\sum Acc_k.
\label{uar}
\end{equation}
where,
\begin{equation}
\footnotesize
F1_k := \frac{2\cdot\sum_{m=1}^{M}TP_k^{(m)}}{2\cdot\sum_{m=1}^{M}TP_k^{(m)}+\sum_{m=1}^{M}FP_k^{(m)}+\sum_{m=1}^{M}FN_k^{(m)}},
\label{f1}
\end{equation}
and
\begin{equation}
\footnotesize
Acc_k = \frac{\sum_{m=1}^{M}TP_k^{(m)}}{n_k},
\label{acck}
\end{equation}
where $K$ is the number of classes,  $M$ is the number of folds of LOSO,  $n_k$ is the total number of samples in the ground truth of the $k$-th class, and $Acc_k$ is the per-class accuracy scores.
For the $m$-th fold of LOSO by the $k$-th class, $TP_k^{(m)}$,  $FP_k^{(m)}$,  and $FN_k^{(m)}$ are true positives, false positives, and false negatives,  respectively.




\section*{Acknowledgements}
This work was supported in part by the National Natural Science Foundation of China under Grant 61672267, and Grant U1836220, the Postgraduate Research \& Practice Innovation Program of Jiangsu Province under Grant KYCX19\_1616, Jiangsu Specially-Appointed Professor Program (No. 3051107219003), Jiangsu joint research project of Sino-foreign cooperative education platform, the Talent Startup project of NJIT (No. YKJ201982), and Central Fund of Finnish Cultural Foundation.

\bibliography{ref_new}
\bibliographystyle{elsarticle-num}

\vbox{}
\vbox{}
\footnotesize
\noindent
\textbf{Ling Zhou}
received her MS degree in in computer science and technology from the Huazhong University of Science and Technology, Wuhan, China in 2011. She is currently working toward the Ph.D. degree with the School of Computer Science and Communication Engineering, Jiangsu University. Her current research interests include multimedia analysis and micro- expression recognition.

\vbox{}
\noindent
\textbf{Qirong Mao}
received her MS and Ph.D. degree from Jiangsu University, Zhenjiang, P. R. China in 2002 and 2009, both in computer application technology. She is currently a professor of the School of Computer Science and Communication Engineering, Jiangsu University. Her research interests include affective computing, pattern recognition, and multimedia analysis. Her research is supported by the key project of National Science Foundation of China (NSFC), Jiangsu province, and the Education Department of Jiangsu province. She has published over 50 technical articles, some of them in premium journals and conferences such as IEEE Trans. on Multimedia, IEEE CVPR, ACM Multimedia. She is a member of the IEEE and ACM.

\vbox{}
\noindent
\textbf{Xiaohua Huang}
received the B.S. degree in communication engineering from Huaqiao  University, Quanzhou, China in 2006. He received his Ph.D degree in Computer Science and Engineering from University of Oulu, Oulu, Finland in 2014. He was a research assistant in Southeast University since 2006. He had been a senior researcher in the Center for Machine Vision and Signal Analysis at University of Oulu in since 2015. He is currently a professor at Nanjing Institute of Technology, China. He has authored or co-authored more than 40 papers in journals and conferences, and has served as a reviewer for journals and conferences. His current research interests include facial expression recognition, micro-expression analysis, group-level emotion recognition, multi-modal emotion recognition and texture classification. He is a member of the IEEE.

\vbox{}
\noindent
\textbf{Feifei Zhang}
received the Ph.D. degree in JiangSu Univerisity, Zhenjiang, Jiangsu, China, in 2019. She is currently an Assistant Professor in Multimedia Computing Group, National Laboratory of Pattern Recognition, Institute of Automation Chinese Academy of Sciences, Beijing, China. Her current research interests include multimedia analysis, computer vision, deep learning, especially multimedia computing, facial expression recognition and cross-modal image retrieval. She has published over 10 technical articles (as the first author), some of them in premium journals and conferences such as IEEE Trans. on Image Processing, IEEE CVPR, and ACM Multimedia.

\vbox{}
\noindent
\textbf{Zhihong Zhang}
received his BSc degree (1st class Hons.) in computer science from the University of Ulster, UK, in 2009 and the PhD degree in computer science from the University of York, UK, in 2013. He won the K. M. Stott prize for best thesis from the University of York in 2013. He is now an associate professor at the software school of Xiamen University, China. His research interests are wide-reaching but mainly involve the areas of pattern recognition and machine learning, particularly problems involving graphs and networks.

\end{document}